\documentclass[journal]{IEEEtran}
\usepackage{amsmath,amsfonts}
\usepackage{algorithmic}
\usepackage{algorithm}
\usepackage{multirow}
\usepackage{array}
\usepackage[table,xcdraw,dvipsnames]{xcolor}
\usepackage{textcomp}
\usepackage{stfloats}
\usepackage{url}
\usepackage{verbatim}
\usepackage{graphicx}
\usepackage{cite}
\usepackage{amssymb}
\usepackage{booktabs}
\usepackage{color}
\usepackage{subcaption}
\usepackage{xcolor}
\usepackage{tikz}
\usepackage{pgfplots}
\pgfplotsset{compat=1.18}
\usepackage{graphicx}
\usepackage{float}
\usepackage{placeins}
\usepackage[normalem]{ulem}
\useunder{\uline}{\ul}{}

\begin{document}

\title{Equal is Not Always Fair: A New Perspective on Hyperspectral Representation Non-Uniformity}

\author{Wuzhou Quan, Mingqiang Wei, Senior Member, IEEE, and Jinhui Tang, Senior Member, IEEE
  \thanks{
    }
  \thanks{W. Quan and M. Wei are with Nanjing University of Aeronautics and Astronautics, Nanjing, China (e-mail: q.wuzhou@gmail.com; mingqiang.wei@gmail.com).}

  \thanks{J. Tang is with Nanjing Forestry Univeristy, Nanjing, China (e-mail: tangjh@njfu.edu.cn).}
  
  }

\markboth{Journal of \LaTeX\ Class Files,~Vol.~14, No.~8, August~2021}%
{Shell \MakeLowercase{\textit{et al.}}: A Sample Article Using IEEEtran.cls for IEEE Journals}

\IEEEpubid{0000--0000/00\$00.00~\copyright~2021 IEEE}

\maketitle

\begin{abstract}
    Hyperspectral image (HSI) representation is fundamentally challenged by pervasive non-uniformity, where spectral dependencies, spatial continuity, and feature efficiency exhibit complex and often conflicting behaviors.
    Most existing models rely on a unified processing paradigm that assumes homogeneity across dimensions, leading to suboptimal performance and biased representations.
    To address this, we propose FairHyp, a fairness-directed framework that explicitly disentangles and resolves the threefold non-uniformity through cooperative yet specialized modules.
    We introduce a Runge-Kutta-inspired spatial variability adapter to restore spatial coherence under resolution discrepancies,
    a multi-receptive field convolution module with sparse-aware refinement to enhance discriminative features while respecting inherent sparsity,
    and a spectral-context state space model that captures stable and long-range spectral dependencies via bidirectional Mamba scanning and statistical aggregation.
    Unlike one-size-fits-all solutions, FairHyp achieves dimension-specific adaptation while preserving global consistency and mutual reinforcement.
    This design is grounded in the view that non-uniformity arises from the intrinsic structure of HSI representations, rather than any particular task setting.
    To validate this, we apply FairHyp across four representative tasks including classification, denoising, super-resolution, and inpaintin, demonstrating its effectiveness in modeling a shared structural flaw.
    Extensive experiments show that FairHyp consistently outperforms state-of-the-art methods under varied imaging conditions.
    Our findings redefine fairness as a structural necessity in HSI modeling and offer a new paradigm for balancing adaptability, efficiency, and fidelity in high-dimensional vision tasks.
    Our code is available at \url{https://github.com/Wuzhou-Quan/HyperFair}.
\end{abstract}


\section{Introduction}

\IEEEPARstart{H}{yperspectral} imaging (HSI) provides significantly finer spectral resolution than conventional imaging modalities, demonstrating remarkable and irreplaceable functionality across many fields of application, including medical diagnosis~\cite{medicalreview}, environmental monitoring~\cite{environmentreview1,environmentreview2}, modern agriculture~\cite{algriculturereview1,algriculturereview2}, military and security~\cite{militaryreview}, and beyond.
Despite the superior potential, HSI presents unique challenges due to its high-dimensional and complex data structure, requiring specialized processing techniques distinct from traditional images.

One of the fundamental challenges in HSI representation learning is inherent non-uniformity, which arises across spectral, spatial, and feature domains. 
Unlike natural RGB images, which exhibit strong local correlations in both adjacent pixels and spectral bands, HSIs present a more complex structure where correlations vary significantly across different domains.
Specifically, the non-uniformity manifests in three aspects:

\begin{itemize}
    \item \textbf{Spatial Variability}:
    Spatial correlations between adjacent pixels vary across imaging modalities.
    Close-range HSIs (e.g., medical or natural imagery) often exhibit strong local correlations due to limited spatial coverage, whereas remote sensing HSIs, with coarser ground sampling, cover larger areas per pixel that may include multiple endmembers, weakening spatial locality assumptions.
    
    \item \textbf{Feature Sparsity}:
    Due to their inherent low-rank and sparse structure, HSIs exhibit highly uneven feature distributions, which pose challenges for conventional feature extraction methods.
    
    \item \textbf{Spectral Irregularity}:
    Correlations among spectral bands are not strictly determined by wavelength adjacency.
    Non-adjacent bands can be more correlated than neighboring ones due to material, sensor, or environmental factors, challenging models that assume local spectral continuity.
\end{itemize}

\begin{figure}[t]
    \centering
    \centering
    \includegraphics[width=.4\textwidth,keepaspectratio]{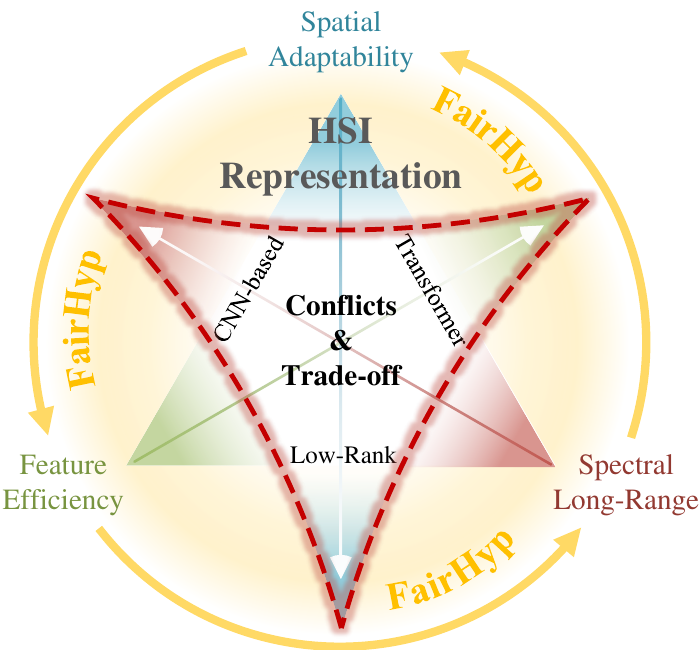}
    \caption{ \label{fig-trilemma}
        Illustration of the trilemma in hyperspectral image (HSI) representation and the motivation of \textbf{FairHyp}.
        The \textcolor[HTML]{8B0000}{red dashed triangle} denotes the trade-offs among Spatial Adaptability, Spectral Long-Range Modeling, and Feature Efficiency, which constrain existing methods.
        \textbf{FairHyp} (shown as the \textcolor[HTML]{BBBB50}{yellow circular path}) resolves this conflict through a modular, fairness-directed design that achieves balanced and adaptive HSI representation.
    }
\end{figure}

\begin{figure*}[!htbp]
    \centering
    \includegraphics[width=\textwidth]{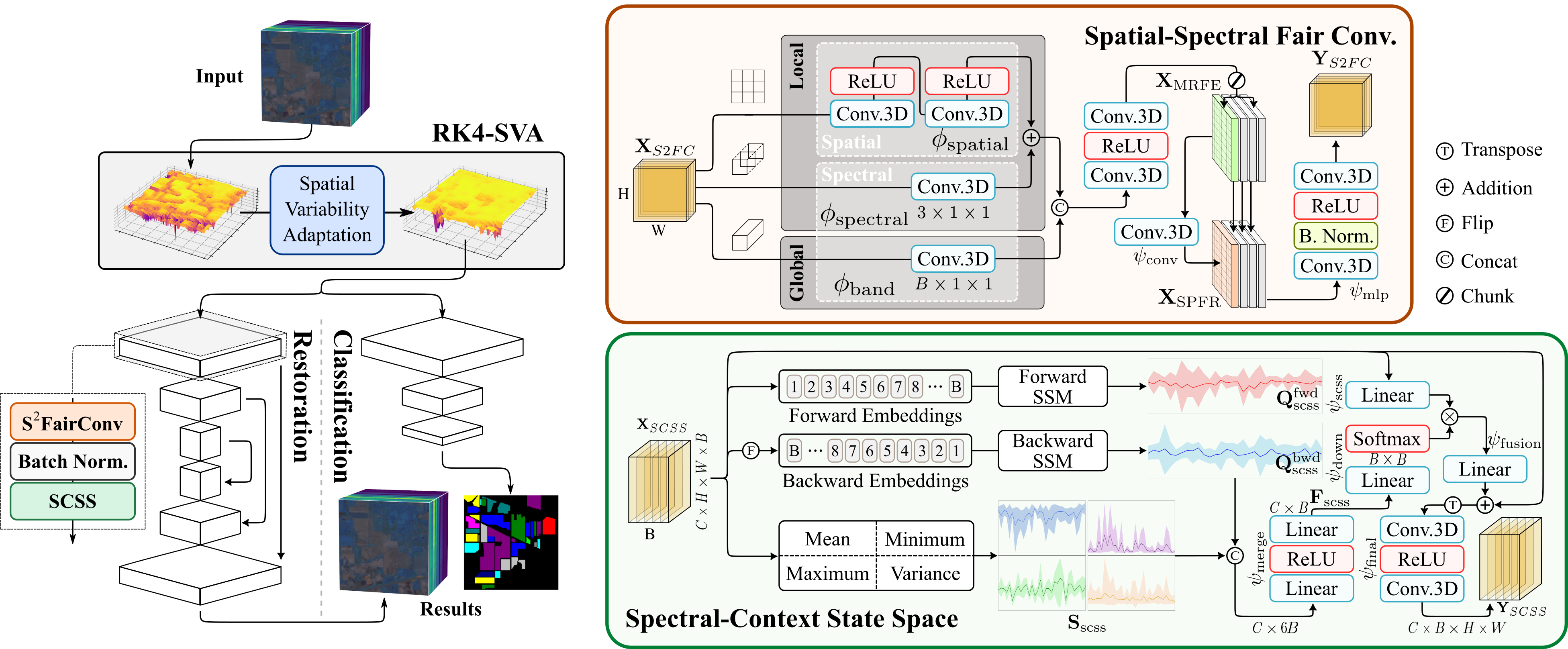}
    \caption{ \label{fig-flow}
        Integration of proposed modules into two representative HSI pipelines.
        RK4-SVA, S$^2$Fair Conv, and SCSS are inserted into restoration (left) and classification (right) networks to address spatial, feature, and spectral non-uniformities, respectively.
        RK4-SVA is detailed separately in Fig.~\ref{fig-rk4sva}.
    }
\end{figure*}

Effectively addressing these issues necessitates the development of adaptive and context-aware models capable of capturing the complex and heterogeneous nature of HSI.
However, most existing methods either focus on a single aspect of non-uniformity~\cite{stcr2023, ng-meet2022, gscvit, drcnn, dstc2024} or apply a uniform adaptation strategy across all dimensions~\cite{9397278,resssnet2024,sert2023,waveformer,SpxDN, HSDT2023}, struggling in the trilemma, as depicted in Fig.\ref{fig-trilemma}.
This equal but unfair solution consequently leads to biased representations and suboptimal performance.
According to Reichstein et al.\cite{review_reichstein2019} and Zhang et al.\cite{review_zhang2022}, physical-driven methods ensure interpretability and generalizability, while data-driven methods provide flexible and adaptive modeling capabilities.
Hence, a hybrid approach that integrates the strengths of both paradigms is essential to overcome the limitations of existing methods.

To this end, we propose \textbf{FairHyp}, a fairness-directed tri-aspect enhancement strategy that explicitly targets the structural flaws underlying HSI representation.
Unlike existing models, FairHyp adopts an independent yet reciprocal design, in which each module is tailored to a specific type of non-uniformity while contributing to a coherent overall solution.
Rather than assuming that a single, unified mechanism could act as a one-size-fits-all solution to all forms of non-uniformity of HSIs, FairHyp adopts a progressive modeling strategy.
This strategy combines model-based priors with data-driven adaptability, enabling the framework to preserve structural commonalities while extracting domain-specific patterns across spectral, spatial, and feature dimensions.
As a result, FairHyp achieves a balanced and context-aware representation that moves beyond the limitations of equal-but-unfair paradigms.
Specifically, FairHyp incorporates three key components:
(1) an adaptive encoding mechanism inspired by the 4th-order Runge-Kutta method (\textit{RK4-SVA}) to harmonize spatial variability across different imaging conditions;
(2) a spatial-spectral fair convolution (\textit{S$^2$Fair}) strategy that mitigates feature sparsity through multi-receptive field extraction and adaptive aggregation;
(3) a spectral-context state space (\textit{SCSS}) model that leverages bidirectional Mamba scanning and statistical modeling to capture long-range spectral dependencies with minimal spatial interference.
Together, these modules form a coherent, interpretable, and fairness-directed enhancement framework for HSI representation.

Through extensive experiments on HSI tasks, including classification denoising, inpainting, and super-resolution, we demonstrate that FairHyp consistently outperforms existing methods by effectively addressing the heterogeneity of HSIs.
These results highlight the robustness and adaptability across diverse applications, reinforcing its role as a principled solution for fair and efficient hyperspectral representation learning.

The contributions of \textbf{FairHyp}, our generic fairness-directed method for HSI representation, are summarized as follows:

\begin{itemize}
\item[1)] We design a spatial variability adapter inspired by the 4th-order Runge-Kutta method to harmonize spatial correlations under diverse imaging conditions.

\item[2)] We introduce a spatial-spectral fair convolution layer that alleviates feature sparsity through multi-receptive field extraction and adaptive feature aggregation.

\item[3)] We develop a spectral-context state space model that leverages Mamba-based bidirectional scanning to capture long-range spectral dependencies while maintaining spectral context coherence.
\end{itemize}

\section{Related Work}

\begin{figure}[t]
    \centering
    \centering
    \includegraphics[width=.45\textwidth,keepaspectratio]{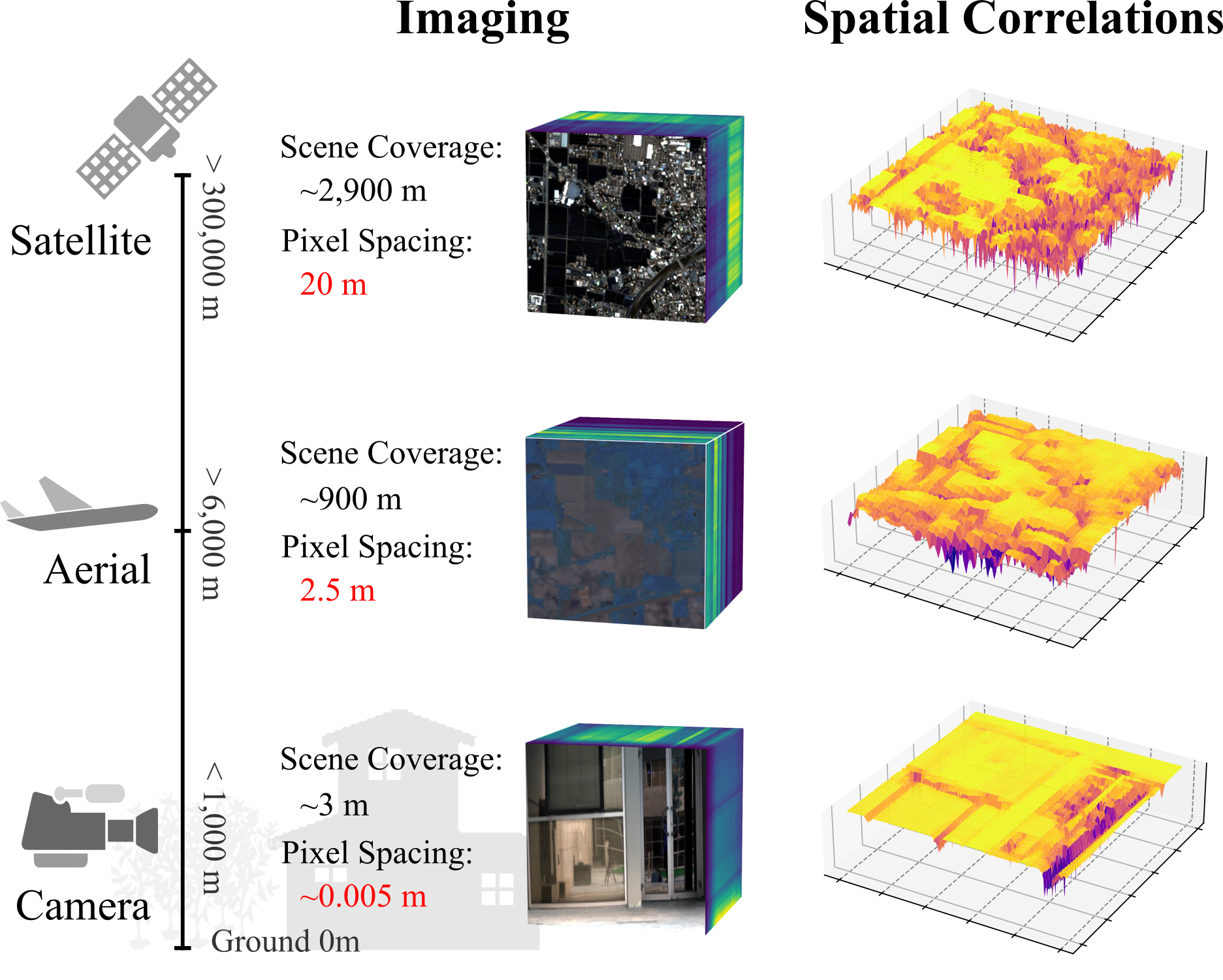}
    \caption{ \label{fig-spatial_variability}
        Spatial variability across imaging sources.
        Different sources produce different pixel spacing, containing fundamentally different spatial content—ranging from mixed-object regions to fine-grained details—thereby affecting spatial correlation patterns.
    }
\end{figure}

HSI analysis has witnessed rapid progress with the adoption of deep learning techniques, enabling more effective modeling of its high-dimensional and complex data structure. 
Among the wide range of tasks in HSI, restoration (e.g., denoising, spatial super-resolution) and classification have become central benchmarks for evaluating representation quality. 
Despite promising advancements, most existing approaches address HSI representation from isolated perspectives, overlooking the intertwined and non-uniform nature of spatial, spectral, and feature dimensions. 
To better contextualize our proposed FairHyp framework, we review recent efforts from the lens of three fundamental challenges in HSI: spatial variability, feature sparsity, and spectral irregularity.

\subsection{Spatial Variability in HSI}

Spatial variability originates from differences in observation distance, spatial resolution, and imaging modalities~\cite{unmixing_review1,unmixing_review2,evidence_sva1}, leading to inconsistent spatial correlations across scenes. 
This discrepancy is particularly evident between near-field and remote sensing scenarios, where the definition of spatial proximity varies significantly.  
Initial efforts to model spatial patterns relied on 3D convolutions~\cite{qrnn3d2021} or combined 2D and 3D structures~\cite{ercsr2021,sfcsr}, which improved multi-scale perception but lacked adaptability to spatial inconsistency.  
To enhance spatial selectivity, attention mechanisms were introduced.  
Han et al.~\cite{unmixing2022} incorporated LiDAR-guided attention to emphasize spatially discriminative regions, while S2VNet~\cite{s2vnet2025} captured subpixel variability using nonlinear encoding and multi-branch fusion.  
Other works applied attention within the HSI itself: Yao et al.~\cite{cscanet2025} used spatial cross-attention for refinement, and He et al.~\cite{lecos2023} applied self-attention to better represent long-range correlations in urban areas.  
More recently, transformer-based architectures have shown advantages in modeling long-range spatial dependencies.  
Ghosh et al.~\cite{transformer_unmixing} combined convolutional encoders with transformers for nonlocal unmixing.  
Yao et al.~\cite{specat2024} proposed cumulative-attention blocks to integrate spatial information progressively, and Wang et al.~\cite{rfsr2022} introduced feedback mechanisms for iterative refinement.  
In parallel, Hou et al.~\cite{pdenet2022} developed a probabilistic embedding framework to address spatial uncertainty in complex environments.  
Although these methods were developed for distinct tasks, they reflect a shared interest in improving spatial representation.  
However, many of them either assume fixed spatial patterns or rely on complex architectures.  
In this work, spatial adaptability is addressed through a dedicated module integrated into a fairness-directed framework to accommodate the spatial characteristics of HSI data.  

\subsection{Feature Sparsity and Low-Rank Nature}

Feature sparsity and low-rank characteristics are intrinsic properties of HSIs, arising from the global and local redundancy across spatial and spectral dimensions.
Early efforts, such as Zhao et al.~\cite{lowrank_zhao2015}, jointly employed sparse representation and low-rank constraints to exploit redundancy and improve denoising performance, yet suffered from spectral distortion due to separate handling of spatial and spectral correlations.
Zhu et al.~\cite{dgmap2014} introduced a data-guided sparsity approach by learning a data-guided map to adaptively impose pixel-wise sparsity constraints.
This method addressed the limitation of applying uniform regularization across heterogeneous pixels, showing that adaptive sparsity can better guide spectral bases and improve unmixing quality.
Subsequent approaches adopted more flexible formulations: He et al.~\cite{ng-meet2022} iteratively constructed orthogonal bases with adaptive rank selection;
Sun et al.~\cite{stcr2023} enhanced robustness by jointly modeling multi-mode correlations via progressive tensor decomposition, though it lacked adaptability to heterogeneous feature distributions.
To improve diversity under sparsity, Hou et al.~\cite{resssnet2024} restructured 3D convolutions with diversity-aware regularization, but remained limited by static architectures.
Transformer-based designs such as Liu et al.~\cite{Liu2025} and Chen et al.~\cite{lrsdn2024} incorporated low-rank priors for efficient global modeling, yet still relied on fixed low-rank assumptions.
In contrast, our FairHyp framework introduces the S$^2$Fair module, which adaptively balances multi-scale spatial-spectral features through receptive field variation and aggregation, moving beyond handcrafted priors to achieve a fairness-directed solution to sparse and uneven HSI representations.

\subsection{Spectral Irregularity and Long-Range Dependencies}

Modeling spectral dependencies in HSIs is particularly challenging due to irregular inter-band correlations.
Unlike RGB images, correlation between adjacent spectral bands is not guaranteed, and distant bands may exhibit strong coupling.
Early efforts, such as non-local low-rank models~\cite{ng-meet2022}, imposed global low-rank constraints on spectral subspaces, achieving compact representation but lacking adaptability to spectral variation.
Convolutional approaches~\cite{hsi-denet2019, 3dadcnn2019, sspn2020} improve local spectral extraction but are limited by their fixed receptive fields and inability to capture global spectral patterns.
To address this, Wei et al.~\cite{qrnn3d2021} proposed a 3D quasi-recurrent pooling strategy to enhance spectral correlation, though it assumes static spectral dependencies.
Sun et al.~\cite{ssftt2022} combined CNNs with transformers to capture both local and global interactions, yet their reliance on spatial-based tokenization introduces spatial bias into spectral modeling.
Transformers have become increasingly popular due to their capacity for long-range modeling via self-attention.
However, the quadratic complexity of self-attention limits scalability to high-resolution HSI data.
To reduce computational costs, Li et al.~\cite{sert2023} integrated lightweight spectral enhancement modules for efficient dependency learning.
Lai et al.~\cite{HSDT2023} designed guided spectral self-attention to improve adaptiveness while maintaining efficiency.
Yao et al.~\cite{specat2024} further refined spectral learning by enforcing dual-domain consistency and using cumulative attention.
Despite these advances, many transformer-based methods intertwine spatial and spectral features during learning, making spectral representations susceptible to spatial variability.
Liu et al.~\cite{dstc2024} employs clustering-based spectral modeling, addressing this by pre-grouping similar spectral patterns to stabilize classification.
However, such task-specific strategies may not generalize well to other HSI applications.
Overall, while recent methods improve spectral modeling through attention and structural priors, most fail to decouple spectral learning from spatial interference.
To address this, our FairHyp introduces a spectral-context state space (SCSS) model that dynamically learns spectral dependencies in both directions using selective scanning and statistical modeling, achieving a fairer and more robust spectral representation.

\textbf{In summary}, while numerous methods have advanced the state-of-the-art in spatial modeling, sparse representation, and spectral dependency learning, they often adopt uniform or task-specific strategies that fail to holistically address the heterogeneous nature of HSI data. 
Such limitations lead to biased or suboptimal representations, especially under complex imaging conditions. 
Our FairHyp framework is motivated by these gaps, proposing a fairness-directed design that explicitly disentangles and adapts to each type of non-uniformity through specialized yet synergistic modules. 
By integrating physical priors with data-driven flexibility, FairHyp offers a unified solution that balances spatial adaptability, feature efficiency, and spectral coherence across diverse HSI tasks.

\section{Method} \label{sec:method}

\subsection{Overview \& Motivation}
\begin{figure}[t]
    \centering
    \begin{subfigure}{0.235\textwidth}
        \centering
        \caption{PaviaU}
        \begin{subfigure}[b]{0.48\textwidth}
            \centering
            \includegraphics[width=\textwidth]{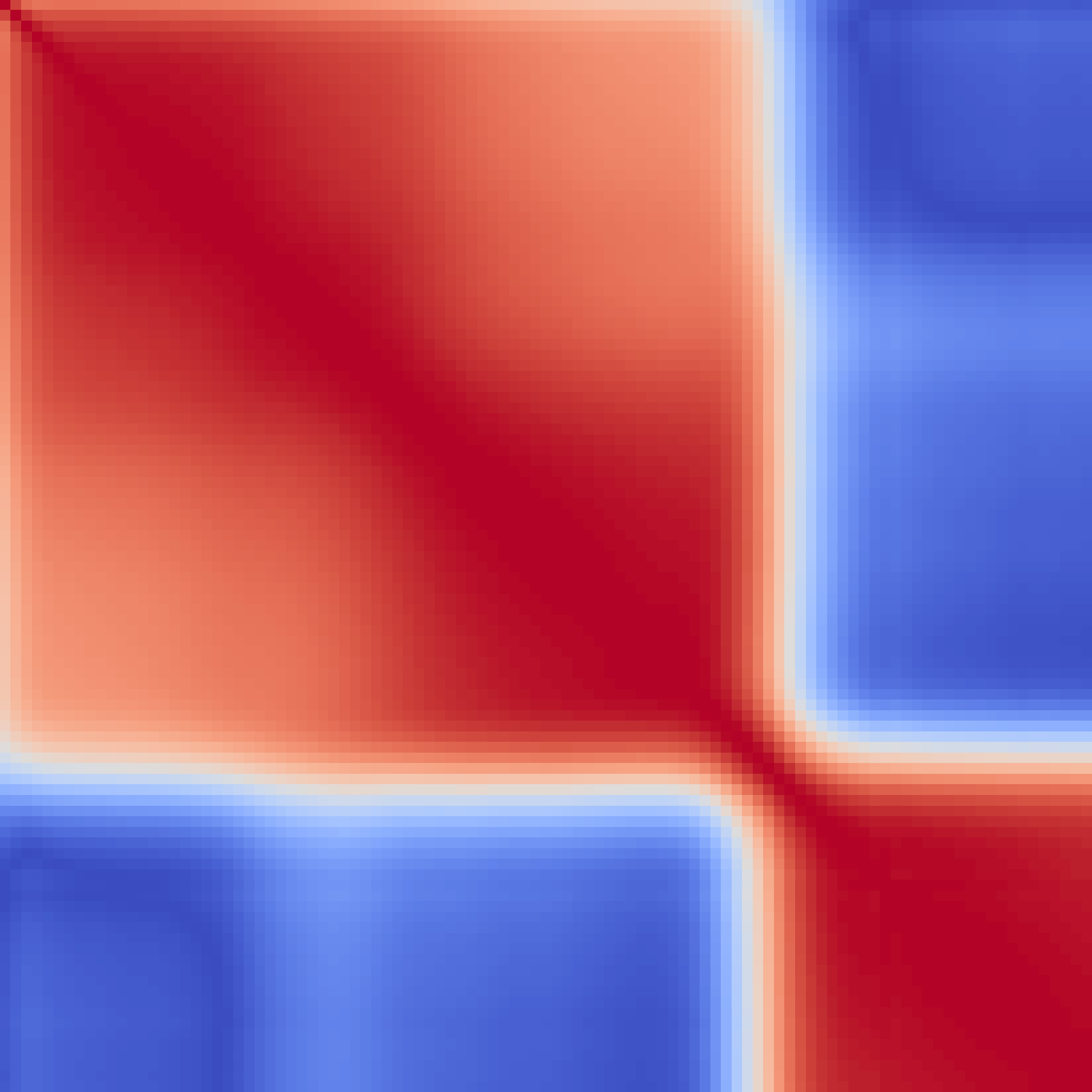}
        \end{subfigure}
        \hfill
        \begin{subfigure}[b]{0.48\textwidth}
            \centering
            \includegraphics[width=\textwidth]{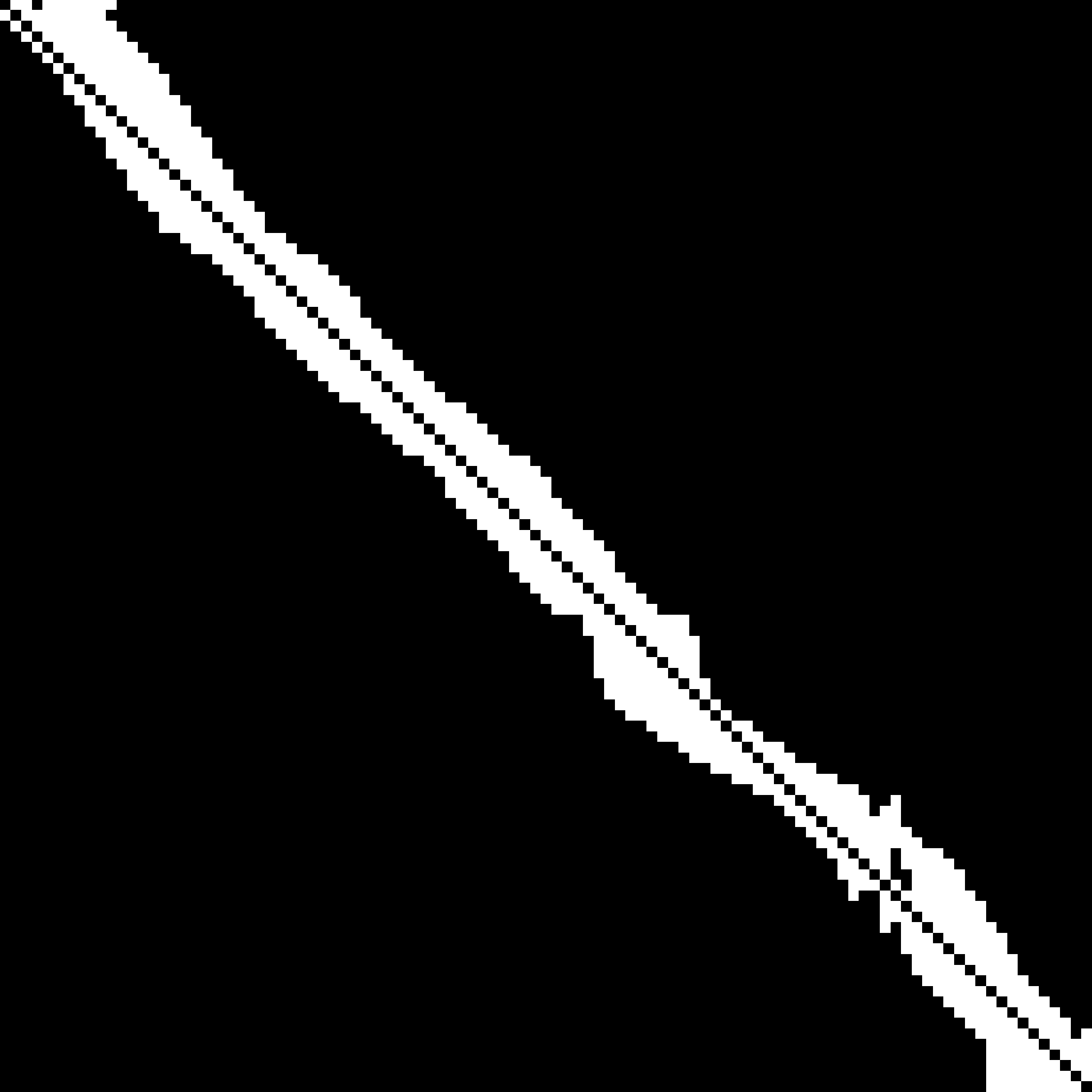}
        \end{subfigure}
    \end{subfigure}
    \hfill
    \begin{subfigure}{0.235\textwidth}
        \centering
        \caption{Indian Pines}
        \begin{subfigure}[b]{0.48\textwidth}
            \centering
            \includegraphics[width=\textwidth]{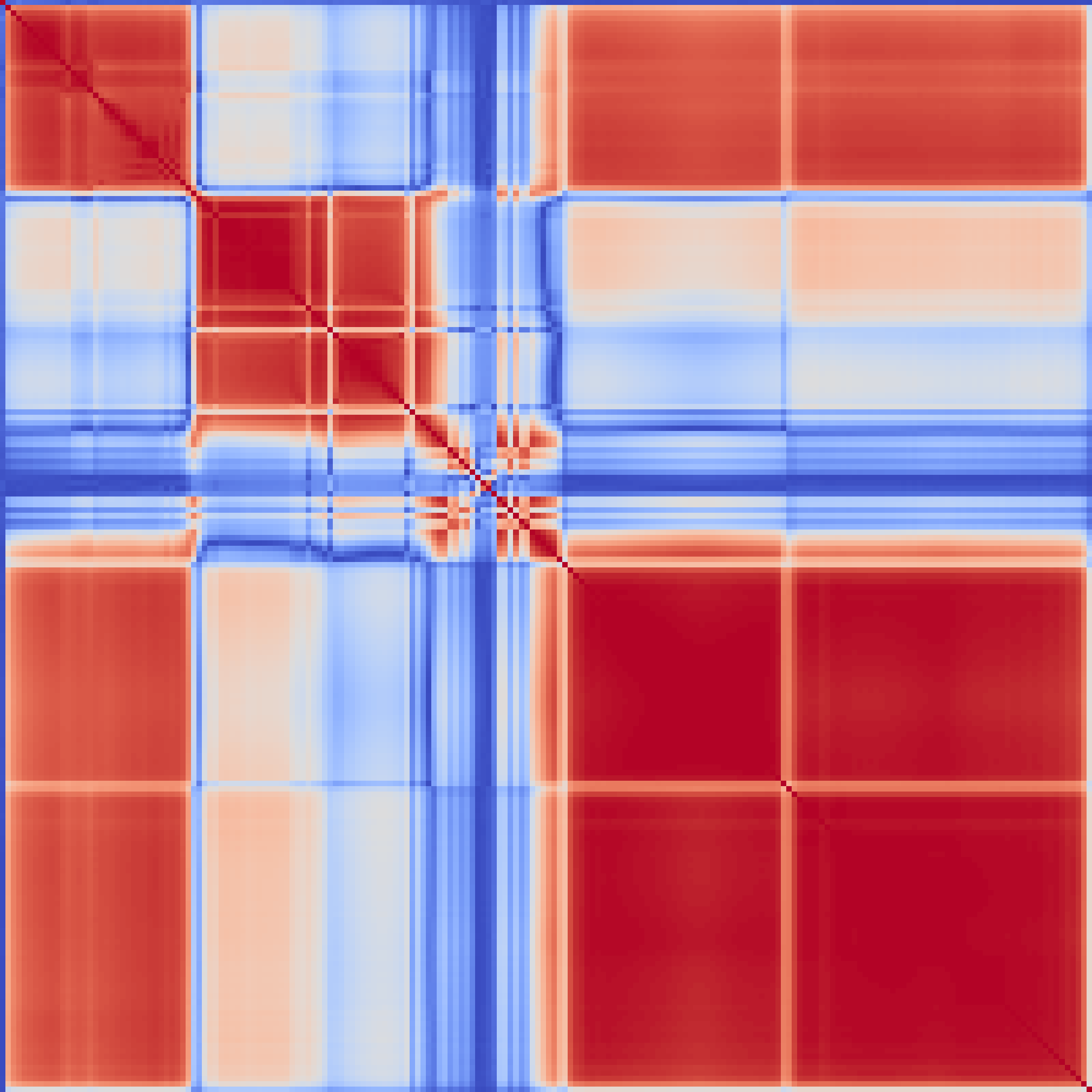}
        \end{subfigure}
        \hfill
        \begin{subfigure}[b]{0.48\textwidth}
            \centering
            \includegraphics[width=\textwidth]{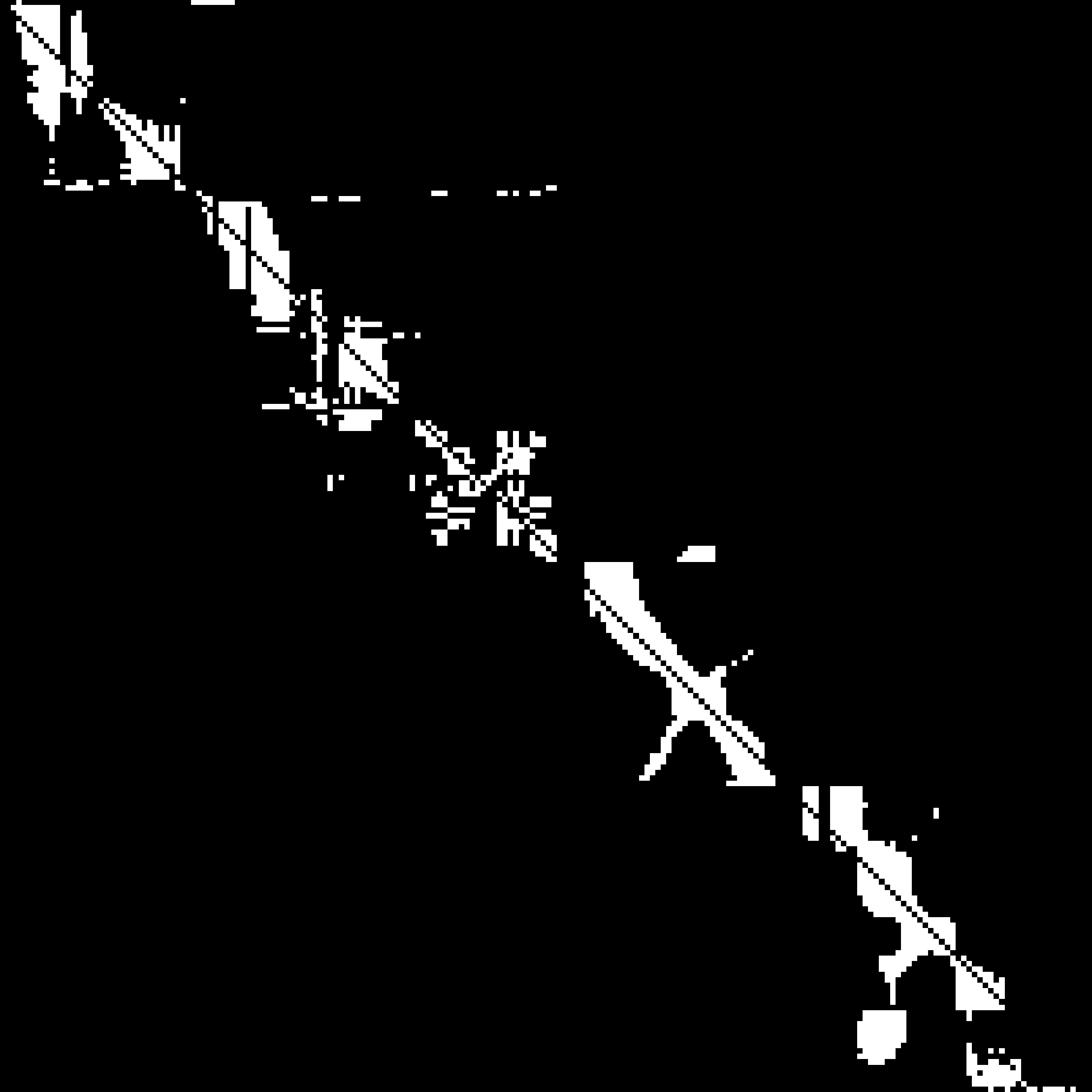}
        \end{subfigure}
    \end{subfigure}
    
    \begin{subfigure}{0.48\textwidth}
        \centering
        \begin{subfigure}[b]{\textwidth}
            \centering
            \caption{Top-k related band average distance of \textit{PaviaU}.}
            \includegraphics[width=\textwidth]{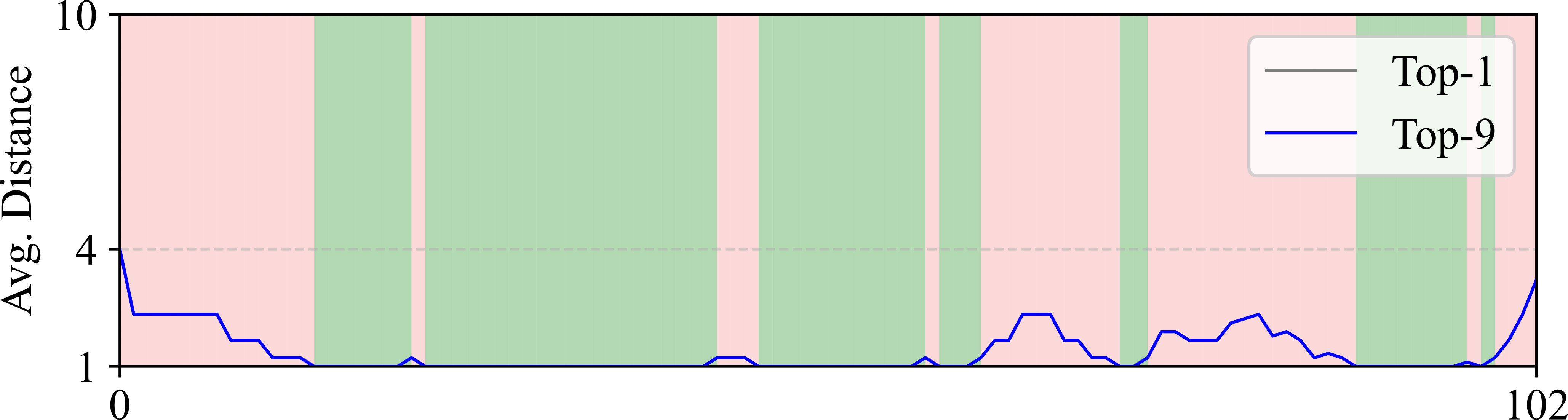}
        \end{subfigure}
        
        \vspace{-1mm}
        \begin{subfigure}[b]{\textwidth}
            \centering
            \caption{Top-k related band average distance of \textit{Indian Pines}.}
            \includegraphics[width=\textwidth]{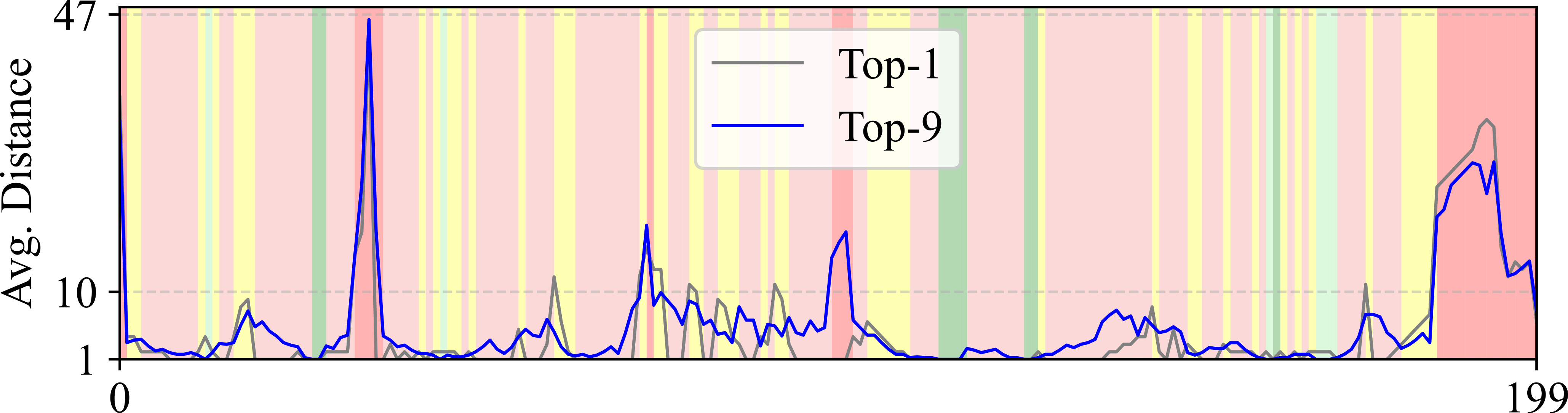}
        \end{subfigure}
    \end{subfigure}

    \vspace{-1.5mm}
    \caption{
        Visualization revealing the breakdown of adjacency-based correlation assumptions in HSIs.
        (a) Band correlation heatmap of the Indian Pines dataset.
        (b) and (c) Top-k related band average distances for Pavia University and Indian Pines, respectively.
        Background colors indicate the adequacy of different receptive fields, as detailed in the text.
    }
    \label{fig-spectral_motivation}
\end{figure}

We design three modules to address the core sources of non-uniformity: \textit{RK4-SVA} for spatial variability, \textit{S}$^2$\textit{FairConv} for feature sparsity, and \textit{SCSS} for spectral inconsistency.
To verify the effectiveness of our design, as illustrated in Fig.~\ref{fig-flow}, we implement these modules across two representative HSI task pipelines: restoration (including spatial super-resolution, inpainting, and real-world denoising) and classification.
Each module is integrated at a distinct stage of the pipeline:
\textit{RK4-SVA} at early encoding to enhance spatial coherence;
\textit{S}$^2$\textit{FairConv} within feature extraction blocks to mitigate sparsity;
and \textit{SCSS} at deeper layers to model long-range spectral dependencies while preserving global context.
Such deliberate decoupling is not arbitrary, but stems from a fairness-directed principle: each form of non-uniformity deserves a tailored solution rather than being uniformly addressed by a monolithic module.
This reflects our belief that fairness in representation learning arises from context-aware adaptation, not from treating inherently different challenges as equal.

\subsubsection{RK4-SVA} \label{sec-3.1.1}

In HSI, spatial heterogeneity arises from varying observation distances and resolutions.
As shown in Fig.~\ref{fig-spatial_variability}, this results in non-uniform adjacent-pixel connectivity, particularly at coarse resolutions, where mixed-object aggregation weakens spatial coherence.
This degradation is most evident near structural boundaries, where the occupancy of a single pixel changes rapidly with resolution, leading to fragmented spatial coherence.
Such effects undermine the locality assumption in convolutional networks and cause sub-pixel information to be severely diluted.
Notably, prior studies have demonstrated the effectiveness of numerical approximations in capturing fine-grained spatial structures.
For example, Zhang et al.~\cite{isnet} employs an explicit Taylor finite-difference scheme to enhance precise edge contrast in infrared images, showing that structural degradations can be addressed by local numerical modeling.
Zhang et al.~\cite{dualode2024} further extend this idea by leveraging ordinary differential equations (ODEs) to model spatial information flow, showing that continuous-depth formulations can improve the recovery of spatial structures under resolution degradation.
Motivated by this, we treat resolution-induced spatial degradation as a degenerate diffusion process and formulate its inversion as a dynamic reconstruction problem.
To solve this, we introduce \textit{RK4-SVA}, an adaptive encoder based on the 4th-order Runge-Kutta method, which progressively restores latent sub-pixel structures from degraded spatial inputs.

\subsubsection{S$^2$FairConv} \label{sec-3.1.2}

Unlike general image processing assumptions of uniform and dense information, HSI features are inherently low-rank and tend to concentrate on a small subset of bands, leaving others redundant and leading to inefficient computation and poor feature utilization~\cite{ghostnet,split2beslim}.
To mitigate this sparsity, previous methods either increase the network width or employ rank reduction techniques.
However, they either increase computational cost or compromise fine-grained structural fidelity.
Inspired by FasterNet~\cite{fasternet}, which decouples computation and memory access by leveraging feature redundancy, we propose the S$^2$Fair convolution module (S$^2$FairConv).
Unlike decomposed or group convolutions~\cite{resssnet2024}, S$^2$FairConv selectively enhances information-rich regions while skipping redundant ones through adaptive partial spatial-spectral convolution.
It processes only a subset of spatial channels, followed by lightweight pointwise aggregation to preserve global consistency.
This design exploits HSI sparsity and avoids over-processing homogeneous or noisy regions.
Multi-scale fields are further used to capture salient structures at varying scales, enabling equitable and efficient feature modeling.
Thus, S$^2$FairConv dynamically allocates capacity where it matters most, achieving fairness through selective attention rather than uniform treatment.

\subsubsection{SCSS}

HSIs exhibit non-uniform long-range spectral dependencies that violate the adjacency assumption commonly adopted by 3D CNNs.  
As shown in Fig.~\ref{fig-spectral_motivation}-(a) and (b), PaviaU shows a clean diagonal in the Pearson correlation heatmap, while Indian Pines displays scattered patterns, indicating disordered correlations.  
To quantify this, we compute the \textit{Top-$k$ Linear-Correlated Band Average Distance} and normalize it against a local adjacency baseline, as visualized in Fig.~\ref{fig-spectral_motivation}-(c) and (d).  
The background is color-coded to reflect model suitability: \colorbox[HTML]{B0FFB0}{green} for local convolutions, \colorbox[HTML]{FFFFB0}{yellow} for moderate receptive field expansion, \colorbox[HTML]{FFD0D0}{light red} for long-range modeling, and \colorbox[HTML]{FFA0A0}{deep red} for global context.  
Indian Pines shows complex, non-adjacent spectral correlations with peak average distances reaching 46—far beyond the receptive field of standard CNNs.  
Some methods~\cite{HSDT2023} have attempted to capture spectral dependencies using multi-head attention, but they are computationally expensive and often unnecessary for datasets like PaviaU, where long-range relationships are minimal.  
The contrast between PaviaU and Indian Pines highlights the complex nature of spectral non-uniformity; prior methods may have over-relied on local modeling that works well on PaviaU, leading to misleading generalizations across diverse HSI datasets.  
These observations motivate the design of the SCSS module, which captures long-range spectral dependencies through bidirectional scanning and statistical aggregation, while maintaining spatial independence.

\subsection{RK4-Inspired Spatial Variability Adapter}

\begin{figure}[t]
    \centering
    \includegraphics[width=.48\textwidth,keepaspectratio]{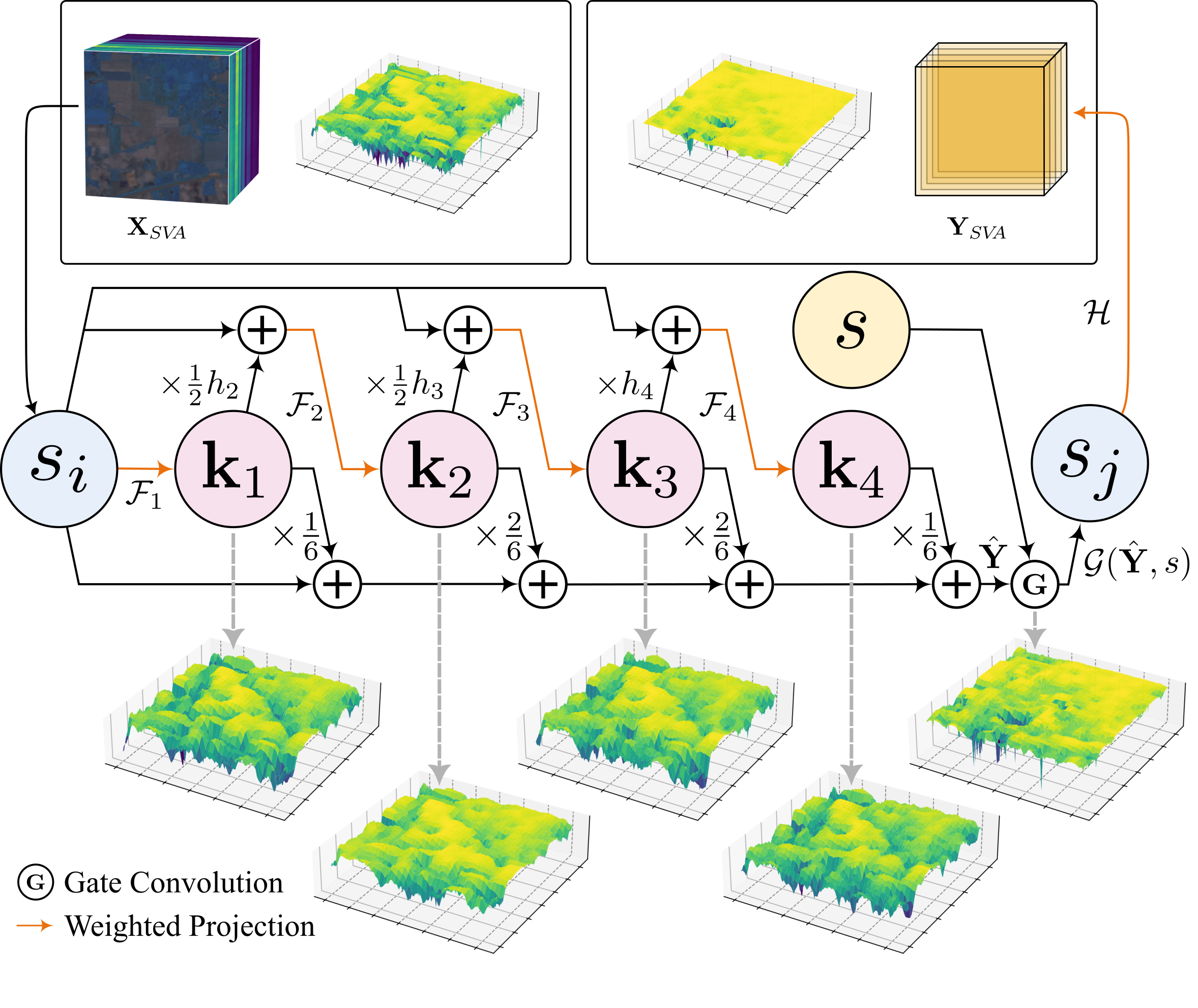}
    
    \caption{ \label{fig-rk4sva}
        Schematic of the RK4-based Spatial Variability Adapter (\textit{RK4-SVA}), which integrates iterative refinement with reference-guided gated modulation for adaptive restoration of spatial correlations.
        The 3D heatmaps illustrate the estimated spatial correlation patterns at different refinement stages.
    }
\end{figure}

To model different spatial structures, we interpret the relationship between discrete pixels and real-world continuous endmembers as the inverse problem of a degenerate diffusion system.
We model the observed spatial signal $\mathbf{X}_{\mathrm{SVA}} \in \mathbb{R}^{B \times H \times W}$ as a degraded state evolved under the dynamics of a first-order ordinary differential equation (ODE):
\begin{equation}
\frac{d\mathbf{X}}{dt} = -\mathcal{F}(\mathbf{X}),
\label{eq:degenerate_ode}
\end{equation}
where $\mathcal{F}$ denotes an unknown degradation function describing the spatial distortion across pixels.
To recover the clean spatial structure, we reformulate the problem as a reconstruction task and solve it using a numerical approximation strategy.
We adopt the classical fourth-order Runge-Kutta (RK4) method, which approximates the solution to initial value problems of the form $\frac{dy}{dt} = f(t, y)$.
Given a current state $y_n$, RK4 computes the next state $y_{n+1}$ by:
\begin{equation}
\begin{aligned}
k_1 &= h f(t_n, y_n), \\
k_2 &= h f\left(t_n + \tfrac{1}{2} h, y_n + \tfrac{1}{2} k_1 \right), \\
k_3 &= h f\left(t_n + \tfrac{1}{2} h, y_n + \tfrac{1}{2} k_2 \right), \\
k_4 &= h f(t_n + h, y_n + k_3), \\
y_{n+1} &= y_n + \tfrac{1}{6}(k_1 + 2k_2 + 2k_3 + k_4),
\end{aligned}
\label{eq:rk4_standard}
\end{equation}
where $h$ is the integration step size.
Inspired by this scheme, we propose a learnable RK4-based spatial adapter, where each \( f(\cdot) \) is replaced by a residual convolutional block that captures local structural variations.  
As illustrated in Fig.~\ref{fig-rk4sva}, the proposed RK4-based adapter consists of four residual steps and a modulation module conditioned on reference features.  
Specifically, we define the update as:
\begin{equation}
\begin{aligned}
\textbf{k}_1 &= \mathcal{F}_{1}(\mathbf{X}_{SVA}), \\
\textbf{k}_2 &= \mathcal{F}_{2}(\mathbf{X}_{SVA} + \tfrac{1}{2} h_2 \cdot \textbf{k}_1), \\
\textbf{k}_3 &= \mathcal{F}_{3}(\mathbf{X}_{SVA} + \tfrac{1}{2} h_3 \cdot \textbf{k}_2), \\
\textbf{k}_4 &= \mathcal{F}_{4}(\mathbf{X}_{SVA} + h_4 \cdot \textbf{k}_3), \\
\hat{\mathbf{Y}} &= \mathbf{X}_t + \tfrac{1}{6}(\textbf{k}_1 + 2\textbf{k}_2 + 2\textbf{k}_3 + \textbf{k}_4),
\end{aligned}
\label{eq:rk4_adapter}
\end{equation}
where each $h_t$ is a learnable scalar controlling the update magnitude.
Each \( \mathcal{F}_t \) represents a separate weighted projection operator, allowing stage-specific adaptation to spatial structures during the process.
To adaptively adjust the reconstructed features under varying imaging conditions, we introduce a reference-aware modulation mechanism.
Specifically, we utilize a reference map \( s \in \mathbb{R}^{C_{\text{ref}} \times H \times W} \), which is derived as a global mapping from the input \( \mathbf{X}_{\mathrm{SVA}} \), to guide spatial adaptation.
The modulation factor \( \alpha \) is computed as:
\begin{equation}
\begin{aligned}
\alpha &= \sigma\left( \mathcal{G}(\hat{\mathbf{Y}}, s) \right), \\
\tilde{\mathbf{Y}} &= \hat{\mathbf{Y}} \cdot (1 + \alpha),
\end{aligned}
\label{eq:ref_modulation}
\end{equation}
where \( \sigma(\cdot) \) denotes the sigmoid function.
Here, \( \mathcal{G} \) denotes a gated spatial convolution defined as:
\begin{equation}
\mathcal{G}(\hat{\mathbf{Y}}, s) = \hat{\mathbf{Y}} \cdot (1 + \alpha), \quad \alpha = \sigma \left( \mathcal{A}([\hat{\mathbf{Y}} \oplus s]) \right),
\label{eq:gated_conv}
\end{equation}
where \( \oplus \) denotes channel-wise concatenation, \( \mathcal{A}(\cdot) \) is a reference-aware modulation generator, and \( \sigma \) denotes the sigmoid activation.
The operator generates a dynamic spatial mask \( \alpha \) conditioned on both the restored features and reference features \( s \)
Finally, a channel-wise refinement operator \( \mathcal{H} \) is applied to yield the output:
\begin{equation}
\mathbf{Y}_{\mathrm{SVA}} = \mathcal{H}(\tilde{\mathbf{Y}}),
\label{eq:channel_refine}
\end{equation}
where \( \mathbf{Y}_{\mathrm{SVA}} \in \mathbb{R}^{B \times H \times W} \) denotes the final adapted representation.
The proposed RK4-SVA thus integrates high-order iterative refinement with reference-conditioned modulation, enabling progressive restoration under non-uniform spatial degradation.

\subsection{Spatial-Spectral Fair Convolution}

Hyperspectral features are often sparse and unevenly distributed across spatial and spectral dimensions, with only a limited subset carrying dominant structure while others remain redundant or noisy.  
Applying uniform convolutions across all regions may lead to biased gradients and unstable representations due to this intrinsic non-uniformity.  
To alleviate this, we propose the Spatial-Spectral Fair Convolution (S$^2$FairConv), which decouples the processing of spatial structure, sparse content, and spectral interaction via a structured and staged design.  
Let $\mathbf{X}_{\text{S2FC}} \in \mathbb{R}^{C \times B \times H \times W}$ denote the input hyperspectral tensor, where $C$ is the channel dimension, $B$ is the number of spectral bands, and $(H, W)$ the spatial resolution.  
The module consists of two stages: multi-receptive field extraction and sparse-preserving refinement.  
An overview of this module is illustrated in the upper-right region of Fig.~\ref{fig-flow}.  

In the first stage, Multi-Receptive Field Extraction (MRFE) decomposes the input through band-sensitive operations:
\begin{equation}
\begin{aligned}
\mathbf{X}_{\text{MRFE}} &= \phi_{\text{spatial}}(\mathbf{X}_{\text{S2FC}}) 
+ \phi_{\text{spectral}}(\mathbf{X}_{\text{S2FC}}) 
+ \phi_{\text{band}}(\mathbf{X}_{\text{S2FC}}).
\end{aligned}
\end{equation}

Here, $\phi_{\text{spatial}}(\cdot)$ applies depthwise convolutions to capture spatial continuity.  
$\phi_{\text{spectral}}(\cdot)$ models inter-band interactions through spectral projections.  
$\phi_{\text{band}}(\cdot)$ encodes global band significance via projection-weighted operations.  
To promote efficient feature separation, only one-fourth of the channels are actively projected in each path, enabling feature skipping and reducing over-processing.  
This decomposition allows for flexible receptive field adaptation across both spatial and spectral dimensions, enriching sparse information with enhanced contextual perception.  
Moreover, the staged strategy enlarges the effective receptive field at lower computational cost compared to conventional dense 3D convolutional layers.  

In the second stage, sparse-preserving feature refinement (SPFR) aims to prevent over-processing of redundant channels:
\begin{equation}
\begin{aligned}
\mathbf{X}_{\text{MRFE}} &= [\mathbf{X}_{\text{act}},\ \mathbf{X}_{\text{passive}}], \\
\mathbf{X}_{\text{conv}} &= \psi_{\text{conv}}(\mathbf{X}_{\text{act}}), \\
\mathbf{X}_{\text{SPFR}} &= \psi_{\text{mlp}}([\mathbf{X}_{\text{conv}}  \oplus \ \mathbf{X}_{\text{passive}}]).
\end{aligned}
\end{equation}

Here, $\psi_{\text{conv}}(\cdot)$ selectively enhances structurally active features through constrained 3D convolutions.  
$\psi_{\text{mlp}}(\cdot)$ integrates updated and passive channels via lightweight projection.  
This selective processing retains low-rank latent information while emphasizing discriminative cues.  
Through this staged decomposition, S$^2$FairConv handles hyperspectral feature sparsity and domain irregularity, producing a balanced and task-adaptive representation.  
It also offers improved representational richness and receptive field coverage, while maintaining computational efficiency by bypassing less-informative channels.

\subsection{Spectral-Context State Space}

HSIs exhibit irregular spectral dependencies, where adjacent bands may not necessarily possess strong correlations, and distant bands can often show unexpected connections.  
Such spectral non-uniformity poses challenges for traditional convolutional or attention-based methods, which typically rely on assumptions of local spectral smoothness.  
To address this issue, we propose the \textbf{Spectral-Context State Space} (SCSS) module, which integrates bidirectional spectral scanning and statistical context aggregation for adaptive spectral modeling.  
The processing pipeline of SCSS is illustrated in the bottom-right part of Fig.~\ref{fig-flow}.  

Let the input tensor be denoted as \( \mathbf{X}_{\text{scss}} \in \mathbb{R}^{C \times B \times H \times W} \), where \(C\) is the number of channels, \(B\) is the number of spectral bands, and \(H\) and \(W\) denote spatial dimensions.  
We first apply forward and backward state-space scanning modules to capture directional spectral context:  
\begin{equation}
\mathbf{Q}^{\text{fwd}}_{\text{scss}} = \text{SSM}_{\text{Forward}}(\mathbf{X}_{\text{scss}}), \quad \mathbf{Q}^{\text{bwd}}_{\text{scss}} = \text{SSM}_{\text{Backward}}(\mathbf{X}_{\text{scss}}),
\end{equation}
where \(\text{SSM}_{\text{Forward}}(\cdot)\) and \(\text{SSM}_{\text{Backward}}(\cdot)\) denote selective spectral scanning in forward and backward directions along the spectral dimension, respectively.  

In parallel, we compute four statistical descriptors across the spatial dimensions to enrich the spectral representation with global context:  
\begin{equation}
\begin{aligned}
\mathbf{S}_{\text{scss}} = \big[ \text{Mean}(\mathbf{X}_{\text{scss}}) \oplus\ & \text{Max}(\mathbf{X}_{\text{scss}}) \\
                                \oplus\ \text{Min}(\mathbf{X}_{\text{scss}}) \oplus\ & \text{Var}(\mathbf{X}_{\text{scss}}) \big],
\end{aligned}
\end{equation}
where \(\text{Mean}(\cdot)\), \(\text{Max}(\cdot)\), \(\text{Min}(\cdot)\), and \(\text{Var}(\cdot)\) compute band-wise statistical summaries over the spatial dimensions.  

The directional embeddings and statistical features are concatenated to form a comprehensive spectral-context representation, which is then refined via a two-layer linear projection:  
\begin{equation}
\mathbf{F}_{\text{scss}} = \psi_{\text{merge}}\left(\left[ \mathbf{Q}^{\text{fwd}}_{\text{scss}} \oplus\ \mathbf{Q}^{\text{bwd}}_{\text{scss}} \oplus\ \mathbf{S}_{\text{scss}} \right]\right).
\end{equation}

To ensure effective spectral modeling with minimal spatial interference, we further apply a residual-enhanced spectral-context fusion:  
\begin{equation}
\mathbf{C}_{\text{scss}} = \psi_{\text{fusion}}\left[\psi_{\text{down}}(\mathbf{F}_{\text{scss}}) \times \psi_{\text{scss}}(\mathbf{X}_{\text{scss}})\right] + \mathbf{X}_{\text{scss}}.
\end{equation}

Finally, a refinement module is applied to produce the output representation:  
\begin{equation}
\mathbf{Y}_{\text{scss}} = \psi_{\text{final}}(\mathbf{F}_{\text{scss}}).
\end{equation}

Through bidirectional spectral scanning, statistical guidance, and adaptive spectral attention, SCSS effectively captures complex spectral dependencies and contributes to fairer and more robust hyperspectral feature representation.

\section{Experiments}

\begin{table*}[t]
    \centering
    
    \caption{Class-wise and overall performance comparison on the \textit{Indian Pines} dataset under a fixed-ratio setting. 
            For each class, 10\% of the available samples are randomly selected as the training set, and the remaining are used for testing. \label{tab:cls_comp_indianpines}}

    \begin{tabular}{c||*{4}{>{\centering\arraybackslash}p{1.2cm}}||
                        *{4}{>{\centering\arraybackslash}p{1.2cm}}||
                        *{2}{>{\centering\arraybackslash}p{1.2cm}}}
    \toprule
         & \multicolumn{4}{c||}{\textbf{CNN-based}} & \multicolumn{4}{c||}{\textbf{ViT-based}} & \multicolumn{2}{c}{\textbf{Ours}} \\
    \midrule
        Class No. & 2DCNN & 3DCNN & CSCANet & ReS$^3$ & ViT & SFormer & SSFTT & DSFormer & FairHyp & FairHyp* \\
    \midrule
        1 & 9.76 & 48.78 & 79.61 & \textbf{95.12} & 17.07 & 80.24 & 78.05 & {\ul 84.54} & 82.93 & 82.93 \\
        2 & 71.60 & 85.37 & \textbf{98.52} & 88.33 & 88.72 & 88.48 & 97.74 & 97.59 & 97.90 & {\ul 98.52} \\
        3 & 41.90 & 76.71 & {\ul 97.32} & 89.56 & 83.80 & 80.05 & 93.84 & \textbf{99.73} & 95.05 & 96.65 \\
        4 & 79.81 & 85.90 & 96.71 & 90.61 & 93.90 & {\ul 98.59} & 96.24 & \textbf{99.53} & 92.02 & 93.43 \\
        5 & 25.35 & 94.24 & 97.93 & {\ul 99.08} & 91.47 & \textbf{99.77} & 96.77 & 93.33 & 97.93 & 97.24 \\
        6 & 86.30 & 98.93 & 98.93 & {\ul 99.54} & 98.48 & 97.85 & 98.78 & 98.33 & 99.39 & \textbf{99.54} \\
        7 & 4.00 & 92.00 & 56.00 & \textbf{100.00} & 24.00 & 72.00 & 88.00 & 73.08 & {\ul 96.00} & \textbf{100.00} \\
        8 & 95.12 & 99.07 & {\ul 99.77} & {\ul 99.77} & 98.60 & 99.30 & \textbf{100.00} & \textbf{100.00} & 99.30 & \textbf{100.00} \\
        9 & 44.44 & 38.89 & 77.78 & \textbf{100.00} & 38.89 & 74.82 & 72.22 & {\ul 83.33} & \textbf{100.00} & 61.11 \\
        10 & 52.75 & 85.01 & 96.66 & 86.73 & 91.42 & 97.71 & \textbf{99.43} & 97.37 & 97.94 & {\ul 98.28} \\
        11 & 87.19 & 90.99 & {\ul 99.09} & 93.62 & 95.07 & 93.93 & 99.05 & 96.47 & \textbf{99.32} & 97.92 \\
        12 & 42.78 & 80.86 & \textbf{98.69} & 87.43 & 86.12 & 92.31 & 94.56 & 97.00 & 94.93 & {\ul 97.19} \\
        13 & 92.39 & 97.29 & \textbf{100.00} & \textbf{100.00} & \textbf{100.00} & \textbf{100.00} & \textbf{100.00} & {\ul 97.84} & \textbf{100.00} & \textbf{100.00} \\
        14 & 96.84 & 96.92 & 98.91 & 98.86 & 98.95 & 96.66 & {\ul 99.47} & 99.30 & 98.07 & \textbf{99.74} \\
        15 & 63.40 & 84.44 & \textbf{98.85} & 96.25 & 94.81 & {\ul 97.98} & 95.68 & 96.84 & 97.12 & 97.12 \\
        16 & 90.36 & 92.77 & 97.59 & \textbf{100.00} & 92.77 & \textbf{100.00} & \textbf{100.00} & 96.43 & 95.68 & {\ul 98.80} \\
        \midrule
        OA~(\%) & 72.51 & 89.22 & 96.73 & 93.40 & 92.51 & 93.27 & {\ul 97.85} & 95.17 & 97.76 & \textbf{98.08} \\
        AA~\%) & 61.50 & 84.26 & 93.27 & {\ul 95.31} & 80.88 & 91.86 & 94.36 & 94.42 & \textbf{96.47} & 94.90 \\
        $\kappa\times10^2$ & 68.22 & 87.68 & 96.27 & 92.47 & 91.44 & 93.12 & {\ul 97.55} & 96.74 & 97.53 & \textbf{97.81} \\
    \bottomrule
    \multicolumn{11}{l}{$^*$The best results are highlighted in \textbf{bold}.
                            The next best results are {\ul underlined}.
                            The same applies to the other tables.}\\
    \end{tabular}
\end{table*}

\subsection{Overall}

To extensively verify the effectiveness of FairHyp, we conduct comprehensive experiments on multiple pixel-level downstream tasks.
Specifically, HSI restoration tasks are essential for ensuring the reliability and effectiveness of HSI data in practical scenarios.
The first category focuses on HSI restoration, which encompasses denoising, blind super-resolution, and blind inpainting.  
These tasks reflect the model’s ability to preserve low-level structural fidelity under diverse degradations, making them ideal for evaluating robustness to spatial variability, spectral irregularity, and feature sparsity.  
To quantify restoration performance, we adopt four widely-used metrics: Peak Signal-to-Noise Ratio (PSNR), Structural Similarity Index Measure~\cite{ssim} (SSIM), Spectral Angle Mapper~\cite{sam} (SAM), and Mean Absolute Error (MAE).  

\begin{itemize}
    \item \textbf{PSNR}: Evaluates the logarithmic ratio between the maximum possible pixel intensity $I_{\max}$ and the mean squared error (MSE) between the reconstructed image $\hat{I}$ and ground-truth image $I$:
    \begin{equation}
        \text{PSNR} = 10 \cdot \log_{10} \left( \frac{I_{\max}^2}{\text{MSE}} \right), \quad \text{MSE} = \frac{1}{N} \sum_{i=1}^{N} (I_i - \hat{I}_i)^2.
    \end{equation}
    
    \item \textbf{SSIM}: Measures perceptual similarity between $I$ and $\hat{I}$ based on luminance, contrast, and structure:
    \begin{equation}
        \text{SSIM}(I, \hat{I}) = \frac{(2\mu_I \mu_{\hat{I}} + C_1)(2\sigma_{I\hat{I}} + C_2)}{(\mu_I^2 + \mu_{\hat{I}}^2 + C_1)(\sigma_I^2 + \sigma_{\hat{I}}^2 + C_2)}.
    \end{equation}
    
    \item \textbf{SAM}: Captures the spectral angle between the true and predicted spectral vectors at each pixel:
    \begin{equation}
        \text{SAM} = \frac{1}{N} \sum_{i=1}^{N} \cos^{-1} \left( \frac{I_i \cdot \hat{I}_i}{\|I_i\| \cdot \|\hat{I}_i\|} \right).
    \end{equation}
    
    \item \textbf{MAE}: Calculates the mean of absolute differences:
    \begin{equation}
        \text{MAE} = \frac{1}{N} \sum_{i=1}^{N} |I_i - \hat{I}_i|.
    \end{equation}
\end{itemize}

The second category involves HSI classification, aimed at evaluating the enhancement in semantic understanding enabled by FairHyp.  
In this context, we employ three standard metrics to assess classification performance: Overall Accuracy (OA), Average Accuracy (AA), and Cohen’s Kappa Coefficient~\cite{kappa} ($\kappa$).

\begin{itemize}
    \item \textbf{Overall Accuracy (OA)}:
    \begin{equation}
        \text{OA} = \frac{\sum_{i=1}^{C} n_{ii}}{N},
    \end{equation}
    where $n_{ii}$ denotes the number of correctly classified samples for class $i$, $C$ is the total number of classes, and $N$ is the total number of test samples.

    \item \textbf{Average Accuracy (AA)}:
    \begin{equation}
        \text{AA} = \frac{1}{C} \sum_{i=1}^{C} \frac{n_{ii}}{n_i},
    \end{equation}
    where $n_i$ is the number of test samples in class $i$.

    \item \textbf{Kappa Coefficient ($\kappa$)}:
    \begin{equation}
        \kappa = \frac{p_o - p_e}{1 - p_e},
    \end{equation}
    where $p_o = \text{OA}$ is the observed agreement and $p_e$ is the expected agreement by chance, computed from the confusion matrix.
\end{itemize}

All experiments are performed on a workstation equipped with a single AMD\textsuperscript{\textregistered} EPYC\texttrademark~7282 CPU and 2 NVIDIA\textsuperscript{\textregistered} RTX\texttrademark 4090 GPUs.  
Experiments are conducted under Ubuntu 22.04 with CUDA 12.1 and PyTorch 2.3.1.

We conduct experiments on four tasks, each with two representative datasets.
In the comparative experiments, we aim to validate the superiority of FairHyp over existing methods.
In the ablation studies, we investigate whether the three modules independently and cooperatively contribute to performance, avoiding feature competition and trade-offs.
Finally, we further explore the functionality, underlying mechanisms, and performance behaviors of each module through additional analyses.

\subsection{HSI Classification} \label{sec:hsic}

\begin{figure*}[t]
    \captionsetup[subfigure]{skip=1pt}
    \centering
    \begin{subfigure}{\textwidth}
        \centering
        \begin{subfigure}[b]{0.195\textwidth}
            \caption{Ground Truth}
            \centering
            \includegraphics[width=\textwidth]{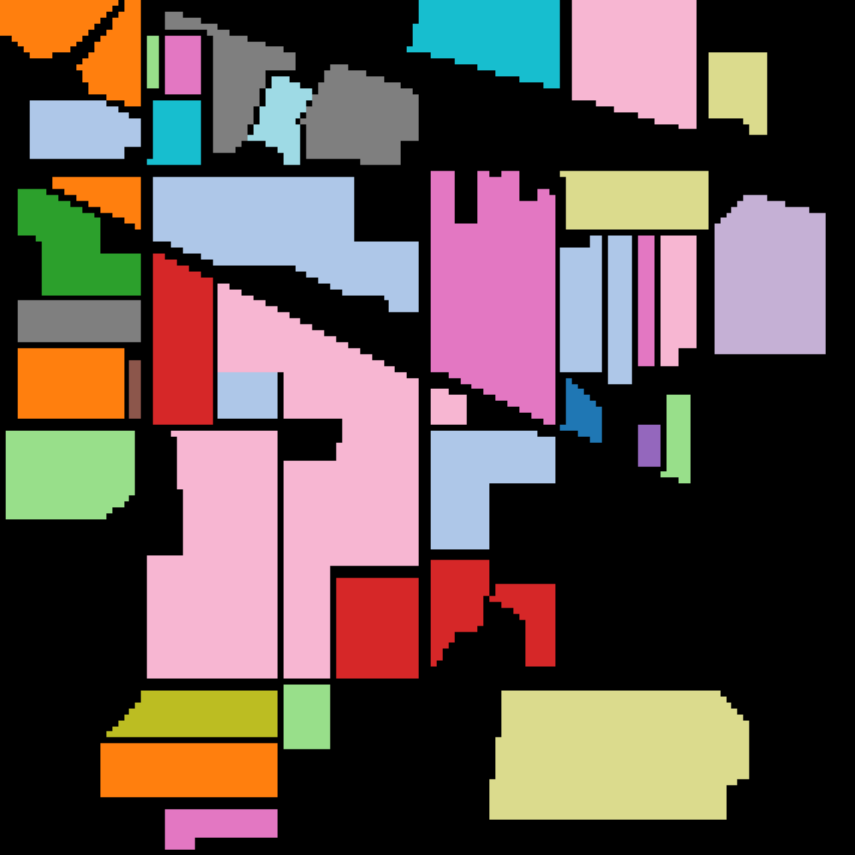}
        \end{subfigure}
        \hfill
        \begin{subfigure}[b]{0.195\textwidth}
            \caption{3DCNN}
            \centering
            \includegraphics[width=\textwidth]{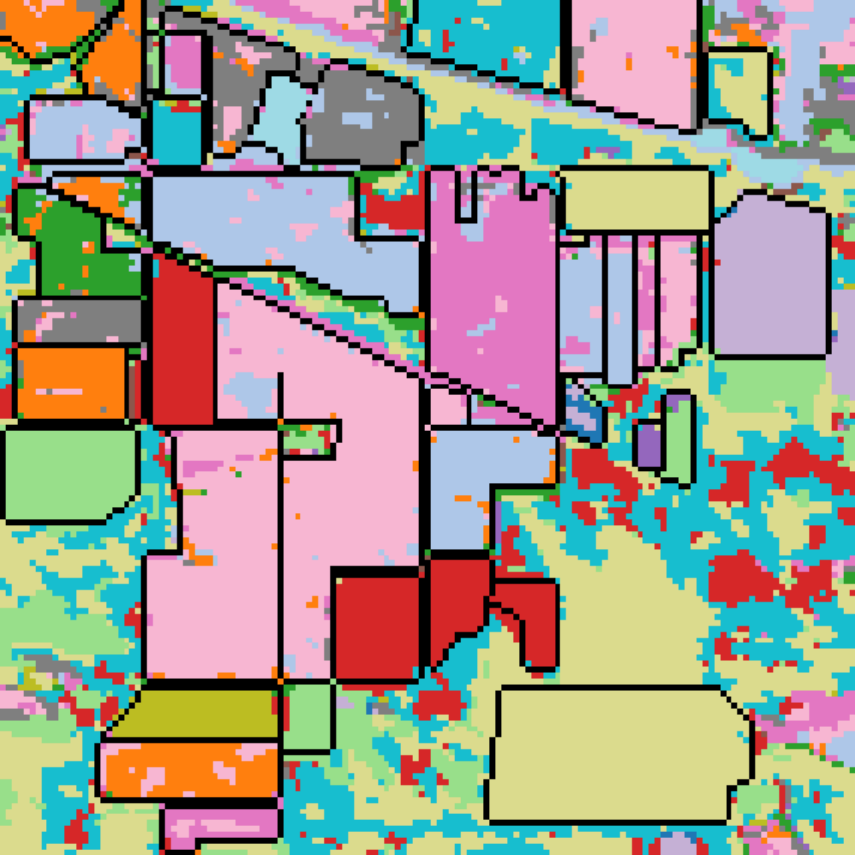}
        \end{subfigure}
        \hfill
        \begin{subfigure}[b]{0.195\textwidth}
            \caption{CSCANet}
            \centering
            \includegraphics[width=\textwidth]{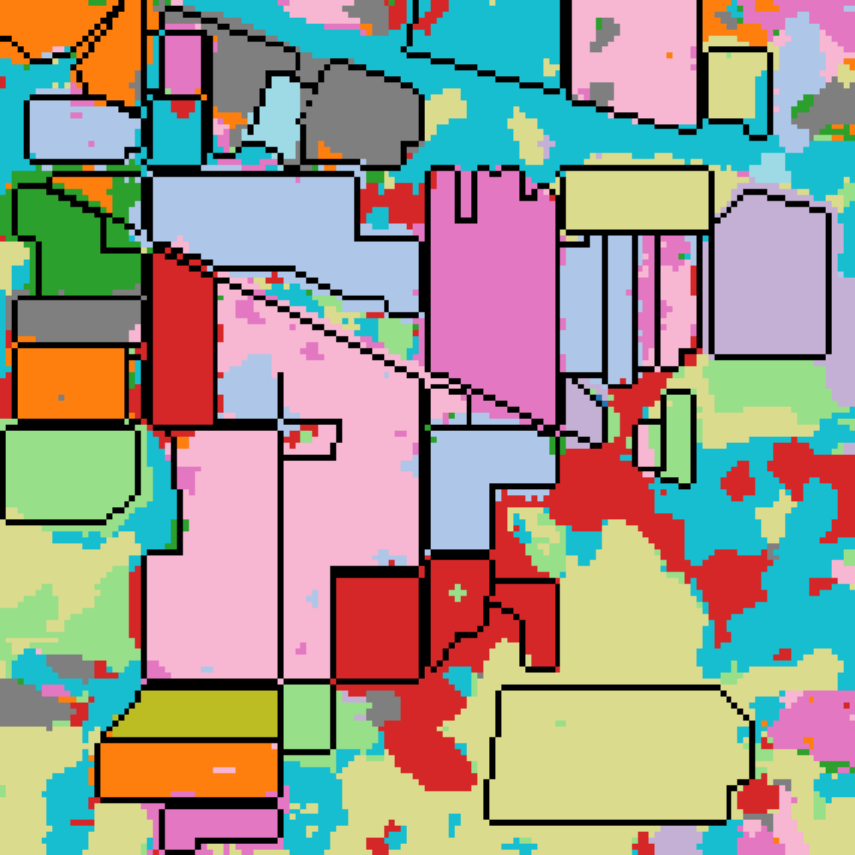}
        \end{subfigure}
        \hfill
        \begin{subfigure}[b]{0.195\textwidth}
            \caption{Res$^3$Conv}
            \centering
            \includegraphics[width=\textwidth]{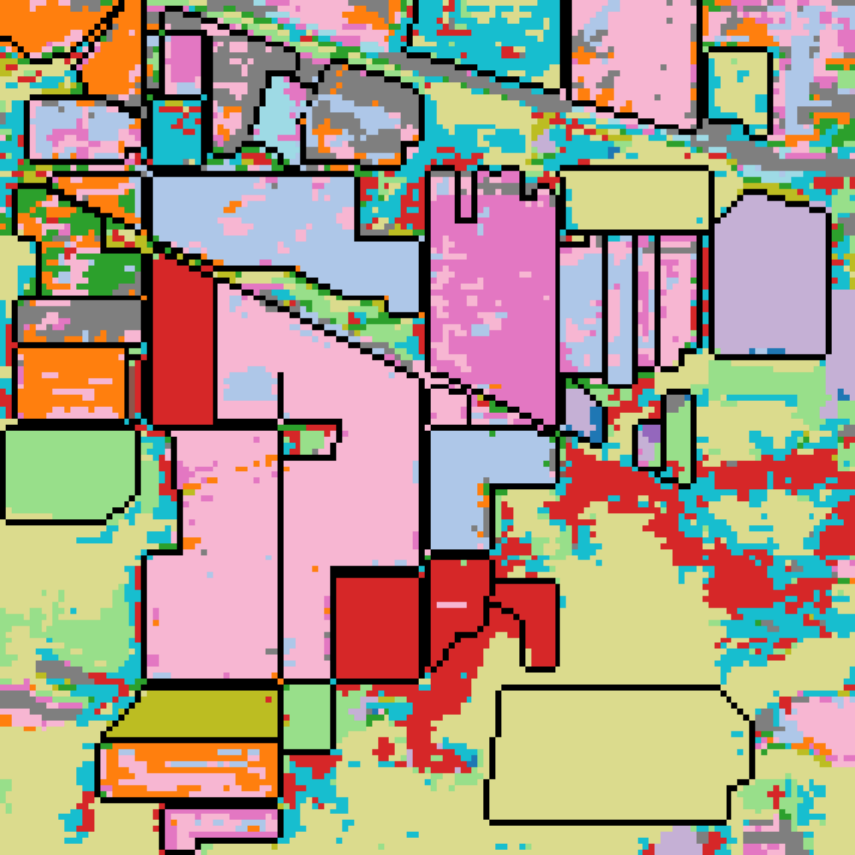}
        \end{subfigure}
        \hfill
        \begin{subfigure}[b]{0.195\textwidth}
            \caption{ViT}
            \centering
            \includegraphics[width=\textwidth]{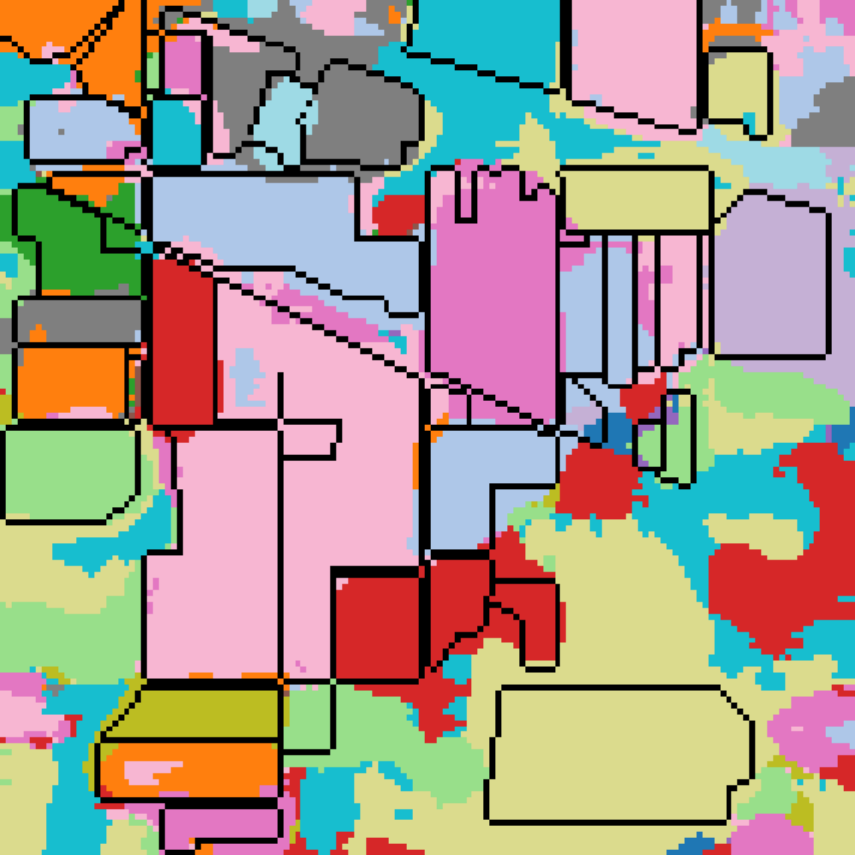}
        \end{subfigure}
    \end{subfigure}
    
    \begin{subfigure}{\textwidth}
        \centering
        \begin{subfigure}[b]{0.195\textwidth}
            \caption{DSFormer}
            \centering
            \includegraphics[width=\textwidth]{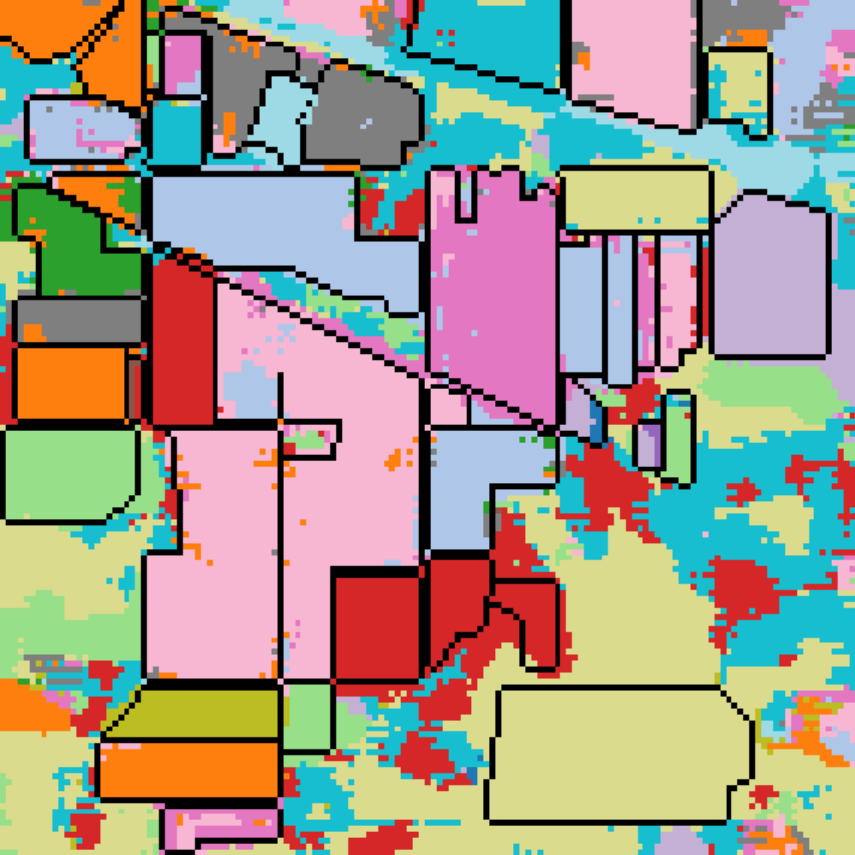}
        \end{subfigure}
        \hfill
        \begin{subfigure}[b]{0.195\textwidth}
            \caption{SpectralFormer}
            \centering
            \includegraphics[width=\textwidth]{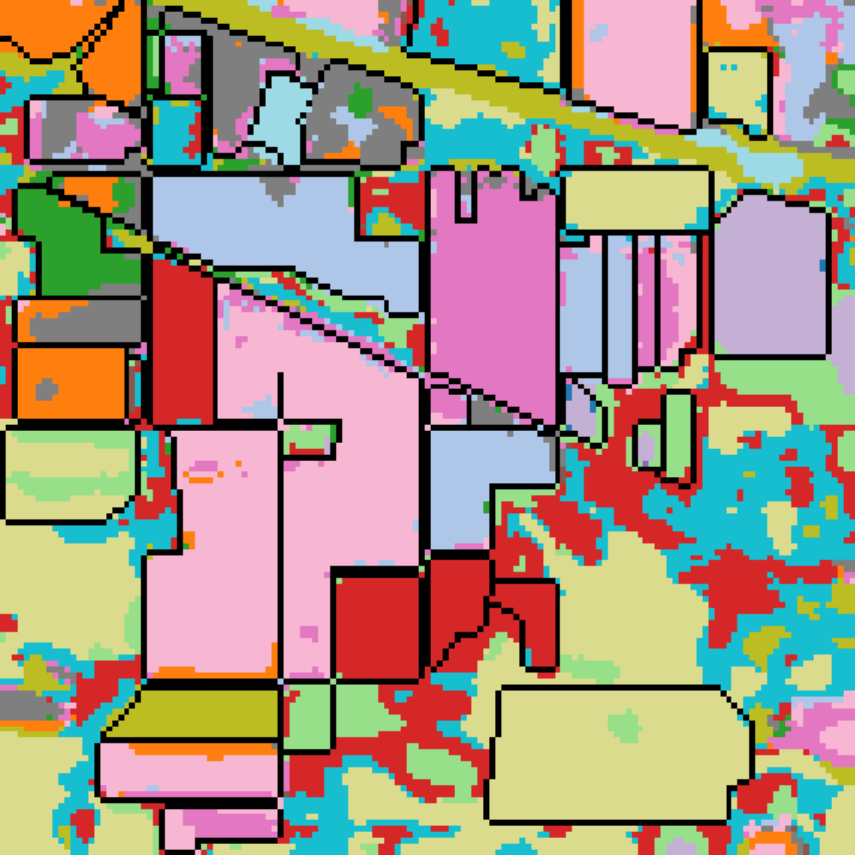}
        \end{subfigure}
        \hfill
        \begin{subfigure}[b]{0.195\textwidth}
            \caption{SSFTT}
            \centering
            \includegraphics[width=\textwidth]{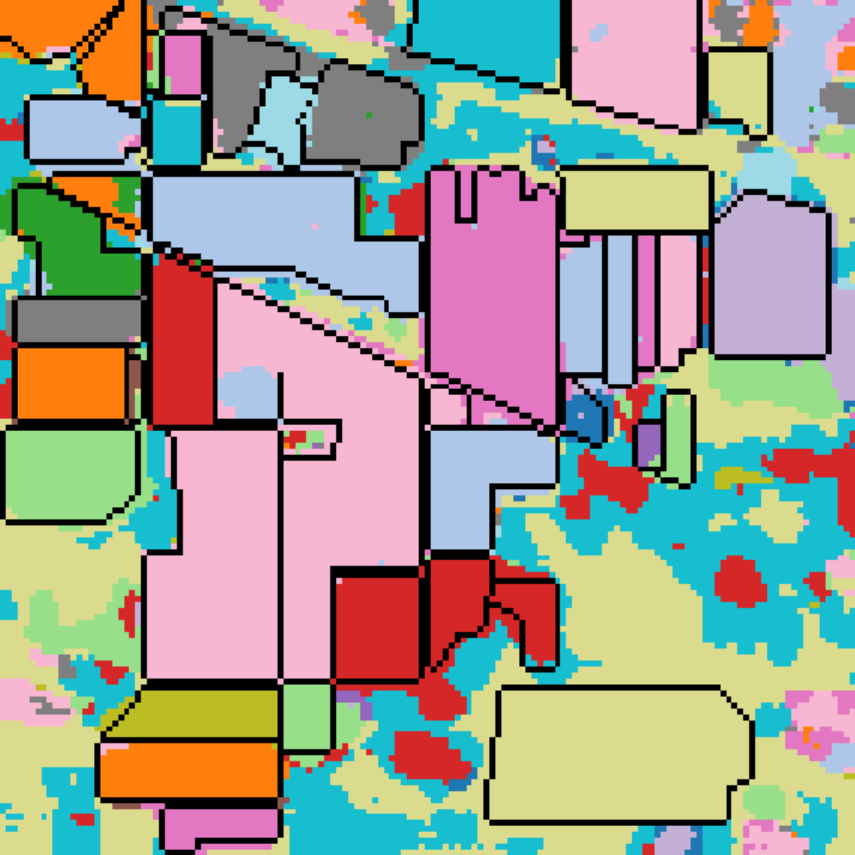}
        \end{subfigure}
        \hfill
        \begin{subfigure}[b]{0.195\textwidth}
            \caption{\textbf{FairHyp} (Ours)}
            \centering
            \includegraphics[width=\textwidth]{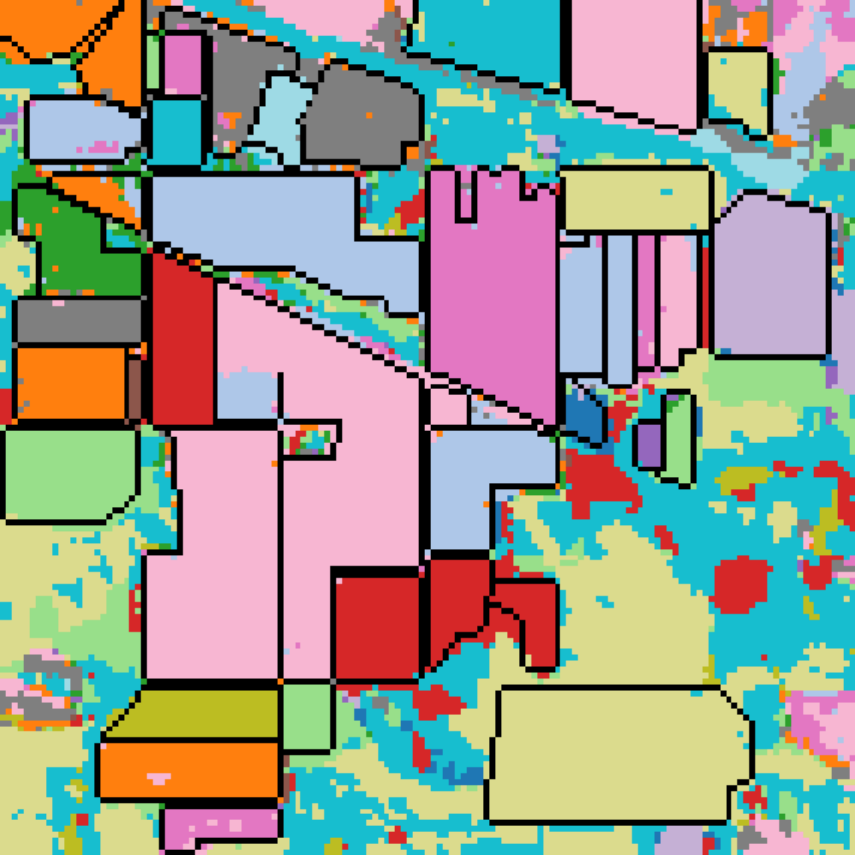}
        \end{subfigure}
        \hfill
        \begin{subfigure}[b]{0.195\textwidth}
            \caption{\textbf{FairHyp*} (Ours)}
            \centering
            \includegraphics[width=\textwidth]{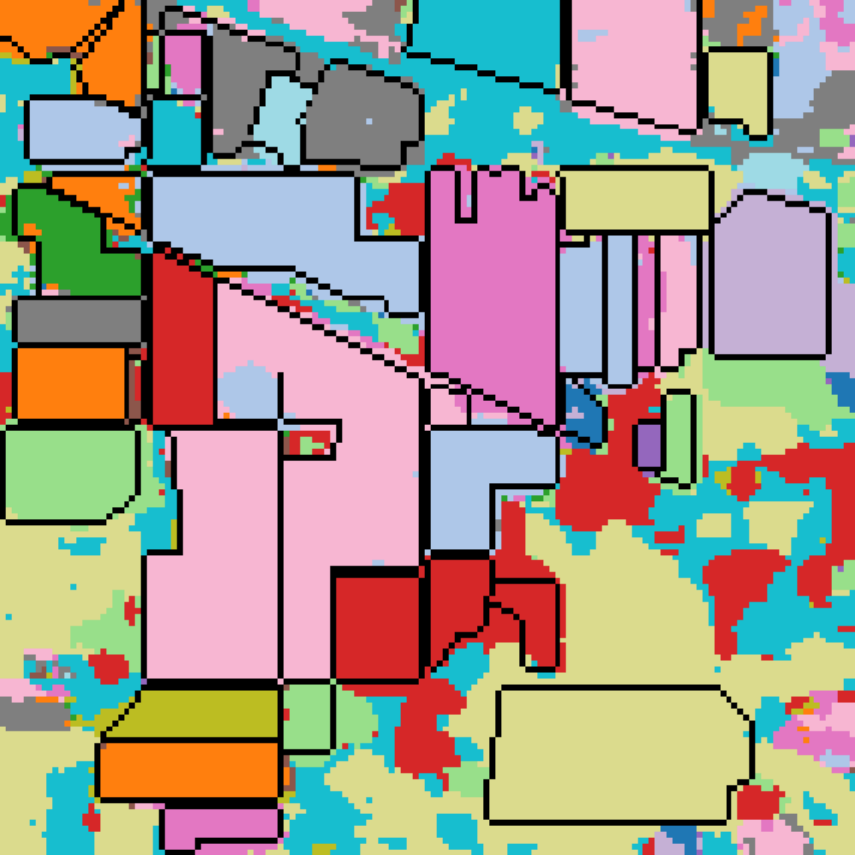}
        \end{subfigure}
    \end{subfigure}
    
    \caption{
        \label{fig:cls_vis_indian_pines}
        Comparisons of the classification visual results of different methods on the \textit{Indian Pines} dataset.
    }
\end{figure*}

\subsubsection{\textbf{Datasets}}
We conduct classification experiments on the Indian Pines\footnote{\url{https://rslab.ut.ac.ir/data}} and WHU-Hi-HongHu\footnote{\url{https://rsidea.whu.edu.cn/resource_WHUHi_sharing.htm}}~\cite{whu-hi2020} datasets.
The Indian Pines dataset, collected by the AVIRIS sensor over northwestern Indiana, consists of $145 \times 145$ pixels and 224 spectral bands (0.4–2.5 µm), covering 16 land-cover classes.
To reduce water absorption and sensor noise effects, we remove 24 bands, retaining 200 bands for experiments.
The WHU-Hi-HongHu dataset was acquired on November 20, 2017, by a Headwall Nano-Hyperspec sensor mounted on a DJI M600 Pro UAV at 100 m altitude in Honghu City, China.
It contains $940 \times 475$ pixels with 270 spectral bands (400–1000 nm) at ~0.043 m spatial resolution.
This dataset features a complex agricultural scene with 23 labeled classes, reflecting diverse crop types and intra-class variability.

\subsubsection{\textbf{Experimental Settings}}
For Indian Pines, we use a fixed‐ratio protocol, selecting 10\% of samples per class for training.
For WHU-Hi-HongHu, we adopt a fixed‐sample protocol, sampling 50 labels per class (0.28\% of all pixels), yielding a per‐class ratio of 1.05\%±1.20\% due to class imbalance.
We train with the AdamW~\cite{adamw} optimizer, an initial learning rate of 1e–4, and cosine annealing from 1e–3 down to 1e–6.
The batch size is 32, and the number of epochs is set to 500.

\subsubsection{\textbf{Comparative Experiments}}
\begin{figure*}[t]
    \centering
    \includegraphics[width=\textwidth]{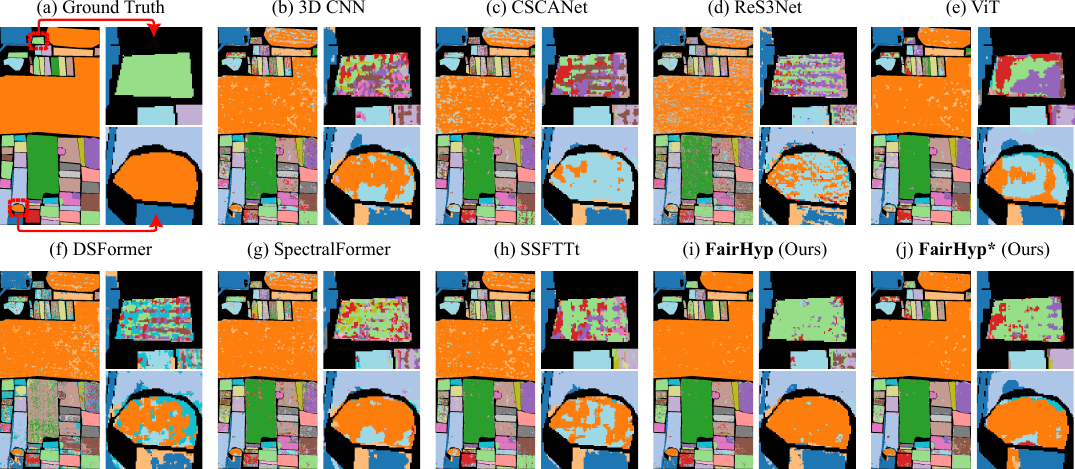}
    \caption{
        \label{fig:cls_vis_honghu}
        Comparisons of the classification visual results of different methods on the \textit{WHU-Hi-HongHu} dataset.
    }
\end{figure*}

We compare our FairHyp with other eight methods including 2D CNN~\cite{cls_sota_2dcnn}, 3D CNN~\cite{cls_sota_3dcnn}, CSCANet~\cite{cscanet2025}, ReS$^3$Conv~\cite{resssnet2024}, ViT~\cite{cls_sota_vit}, Spectral Former~\cite{spectralformer2022}, SSFTT~\cite{ssftt2022}, and DSFormer~\cite{dsformer2025}.
To validate the generalizability of our framework beyond architectural difference in classification tasks, we further implement a Transformer-based variant of FairHyp built upon SSFTT, denoted as FairHyp*.
As shown in Table~\ref{tab:cls_comp_indianpines}, on the Indian Pines dataset, FairHyp* achieves the highest OA (98.08\%) and $\kappa$ (97.81\%), outperforming both CNN and Transformer baselines, and FairHyp achieves the highest AA (96.47\%).
As shown in Table~\ref{tab:cls_comp_honghu}, On the \textit{WHU-Hi-HongHu} dataset, FairHyp* achieves the best OA (93.09\%), AA (94.26\%), and $\kappa$ (92.38\%).
It outperforms others on most categories, benefiting from our tri-aspect enhancement strategy.
FairHyp* shows comparable performance, confirming that our designing enhances both CNN and Transformer designs.
Notably, previous CNN-based methods perform poorly in this scenario, while FairHyp consistently outperforms them across all metrics, ranking just behind FairHyp*.
This suggests that the performance gap may not stem from the architectural type itself, but rather from the lack of fair enhancements in prior designs.
It demonstrates superior robustness in handling small-sample categories and irregular class distributions, mitigating the limitations of limited receptive fields and early tokenization.
Visualization results in Fig.~\ref{fig:cls_vis_indian_pines} and~\ref{fig:cls_vis_honghu} further confirm its advantages.

\begin{table*}[t]
    \centering
    
    \caption{Class-wise and overall performance comparison on the \textit{WHU-Hi-HongHu} dataset under a fixed-sample setting. 
            Each class contains exactly 50 training samples, with the remaining used for testing.\label{tab:cls_comp_honghu}}
    \begin{tabular}{c||*{4}{>{\centering\arraybackslash}p{1.2cm}}||
                        *{4}{>{\centering\arraybackslash}p{1.2cm}}||
                        *{2}{>{\centering\arraybackslash}p{1.2cm}}}
    \toprule
        & \multicolumn{4}{c||}{\textbf{CNN-based}} 
        & \multicolumn{4}{c||}{\textbf{ViT-based}} 
        & \multicolumn{2}{c}{\textbf{Ours}} \\
    \midrule
        Class No. & 2DCNN & 3DCNN & CSCANet & ReS$^3$ & ViT & SFormer & SSFTT & DSFormer & FairHyp & FairHyp* \\
    \midrule
        1 & 90.76 & 92.66 & 95.53 & 85.67 & 95.13 & 93.73 & 95.21 & {\ul 96.13} & 96.10 & \textbf{96.55} \\
        2 & 85.90 & 86.45 & 89.63 & 86.37 & 87.23 & 88.15 & {\ul 96.13} & 94.42 & 89.38 & \textbf{97.47} \\
        3 & 78.52 & 75.76 & 83.29 & 77.16 & 70.73 & 82.20 & 88.51 & 83.47 & \textbf{90.09} & {\ul 89.27} \\
        4 & 86.62 & 88.56 & 76.83 & 81.40 & 93.26 & 91.55 & 89.58 & {\ul 95.08} & \textbf{98.24} & 94.67 \\
        5 & 78.53 & 78.94 & 82.96 & 78.68 & 91.75 & 85.80 & 92.38 & 92.51 & {\ul 92.76} & \textbf{97.84} \\
        6 & 80.92 & 89.27 & 90.83 & 81.53 & 93.88 & 88.30 & 94.68 & 93.26 & \textbf{97.35} & {\ul 96.76} \\
        7 & 54.55 & 60.03 & 49.10 & 62.30 & 68.03 & 63.17 & 67.73 & 71.07 & \textbf{86.15} & {\ul 81.83} \\
        8 & 34.99 & 38.76 & 62.21 & 41.93 & 66.38 & 53.80 & {\ul 76.91} & 68.75 & 76.42 & \textbf{88.51} \\
        9 & 97.25 & 94.14 & 95.12 & 94.08 & 95.17 & 93.89 & 98.09 & 96.75 & {\ul 98.32} & \textbf{98.96} \\
        10 & 71.52 & 64.47 & 74.57 & 66.12 & 86.95 & 56.57 & 87.06 & 72.24 & {\ul 87.12} & \textbf{92.84} \\
        11 & 44.16 & 63.64 & 72.39 & 57.84 & 65.50 & 60.33 & 72.54 & 76.77 & {\ul 79.75} & \textbf{88.82} \\
        12 & 57.95 & 58.87 & 76.68 & 60.24 & 66.24 & 70.20 & 78.01 & 75.46 & {\ul 83.28} & \textbf{88.02} \\
        13 & 63.82 & 63.07 & 76.32 & 62.94 & 77.41 & 66.43 & 81.01 & \textbf{81.57} & 73.50 & {\ul 81.43} \\
        14 & 82.67 & 76.16 & 94.03 & 77.46 & 90.10 & 82.10 & {\ul 95.43} & 92.70 & 92.35 & \textbf{96.56} \\
        15 & 83.93 & 92.33 & 86.45 & 91.18 & 96.74 & 94.41 & {\ul 97.80} & 96.41 & 96.41 & \textbf{99.60} \\
        16 & 89.79 & 83.72 & {\ul 96.16} & 80.26 & 88.69 & 80.90 & 95.40 & 94.19 & 93.90 & \textbf{97.56} \\
        17 & 91.25 & 82.53 & 93.21 & 81.76 & 88.75 & 91.66 & {\ul 97.08} & 90.90 & 92.79 & \textbf{97.67} \\
        18 & 83.86 & 87.09 & 92.71 & 82.38 & 90.62 & 89.37 & 97.39 & 95.55 & {\ul 97.45} & \textbf{98.79} \\
        19 & 80.37 & 84.78 & 90.08 & 84.78 & 88.93 & 88.01 & 90.74 & {\ul 92.77} & 91.85 & \textbf{94.38} \\
        20 & 86.38 & 93.04 & 88.30 & 87.92 & 94.38 & 90.22 & {\ul 97.82} & 97.36 & 95.09 & \textbf{98.22} \\
        21 & 84.19 & 80.05 & {\ul 99.69} & 82.32 & 98.28 & 86.22 & 99.55 & 97.97 & 95.33 & \textbf{99.85} \\
        22 & 66.17 & 88.62 & {\ul 97.34} & 86.77 & 95.74 & 90.52 & 96.09 & 93.64 & 95.87 & \textbf{98.19} \\ \midrule
        OA~(\%) & 79.23 & 81.66 & 79.75 & 77.49 & 87.47 & 84.03 & 88.39 & 89.86 & {\ul 91.33} & \textbf{93.09} \\
        AA~(\%) & 76.10 & 78.32 & 84.70 & 76.87 & 85.90 & 81.25 & 90.23 & 88.59 & {\ul 90.89} & \textbf{94.26} \\
        $\kappa\times10^2$ & 74.55 & 77.42 & 75.69 & 72.73 & 84.41 & 80.22 & 85.67 & 87.34 & {\ul 91.60} & \textbf{92.38} \\
    \bottomrule
    \end{tabular}
\end{table*}

\subsection{HSI Denoising} \label{sec:hsid}

\begin{figure*}[t]
    \centering
    \includegraphics[width=\textwidth]{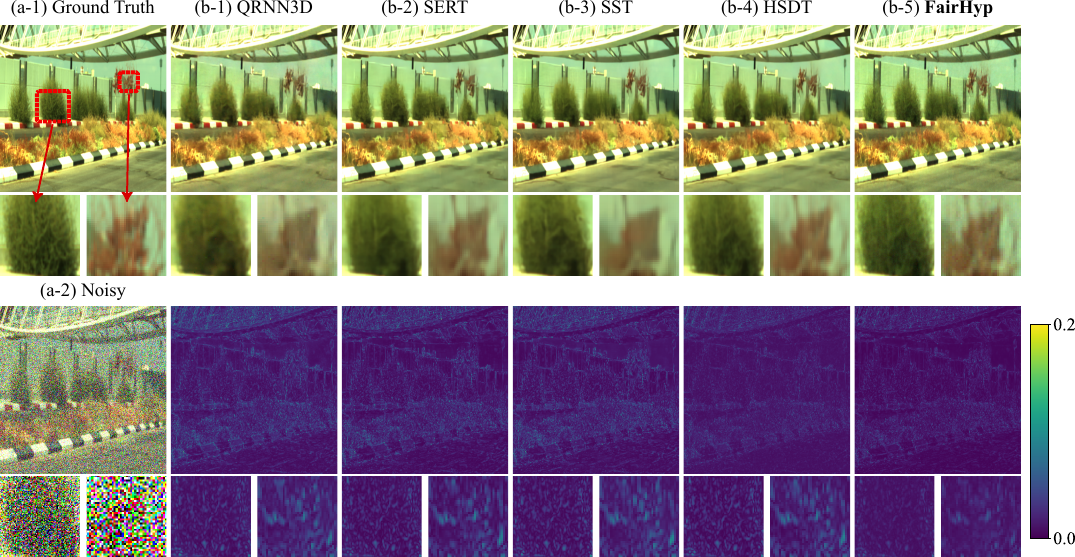}
    \caption{
        \label{fig:dn_vis_icvl}
        Comparisons of the denoising visual results of different methods on the \textit{ICVL} dataset with \textit{Blind Gaussian} noise.
    }
\end{figure*}

\begin{figure*}[t]
    \centering
    \includegraphics[width=\textwidth]{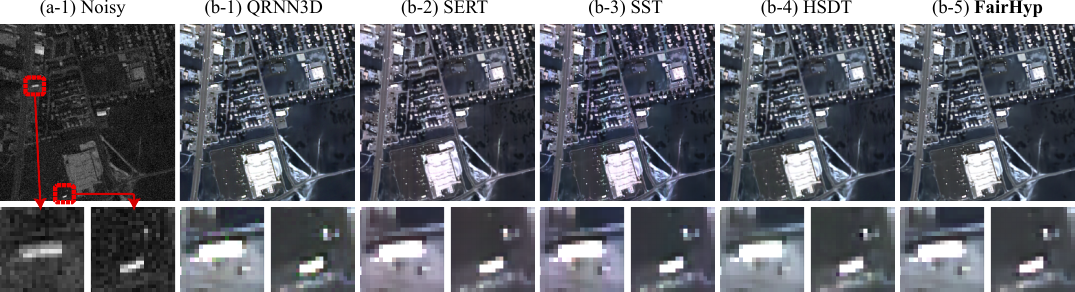}
    \caption{
        \label{fig:dn_vis_urban}
        Comparisons of the denoising visual results of different methods on the \textit{Urban} dataset with \textit{Realistic} noise.
    }
\end{figure*}

\begin{table*}[t]
\centering
\caption{Quantitative comparison of denoising results on the \textit{ICVL} dataset.\label{tab:dn_comp_icvl}} 
\centering
\begin{tabular}{c|*{1}{>{\raggedleft\arraybackslash}p{1.8cm}}||*{1}{>{\centering\arraybackslash}p{1.1cm}}|*{3}{>{\centering\arraybackslash}p{1.1cm}}|*{4}{>{\centering\arraybackslash}p{1.1cm}}|*{1}{>{\centering\arraybackslash}p{1.1cm}}}
\toprule
\multicolumn{3}{c|}{\textbf{}} & \multicolumn{3}{c|}{\textbf{Model Driven}} & \multicolumn{4}{c|}{\textbf{Data Driven}} & \textbf{Ours} \\
\midrule
\multicolumn{2}{c||}{\textbf{Noise Type}} & Noisy & BM4D & LRMR & NGMeet & QRNN3D & SST & SERT & HSDT & \textbf{FairHyp} \\
\midrule
\multirow{4}{*}{\textbf{30}}    & \textbf{PSNR}~(dB)~$\uparrow$  & 18.59 & 30.21 & 32.45 & 41.67 & 41.48 & 41.99 & 42.58 & {\ul 42.81} & \textbf{42.91} \\
                                & \textbf{SSIM}~(\%)~$\uparrow$  & 10.25 & 89.12 & 91.35 & 97.65 & 97.44 & 97.86 & {\ul 97.96} & 97.52 & \textbf{98.23} \\
                                & \textbf{SAM}~(\%)~$\downarrow$ & 78.20 & 18.24 & 15.31 & 5.02 & 6.24 & 5.06 & {\ul 4.73} & 5.36 & \textbf{4.46} \\
                                & \textbf{MAE}~(‰)~$\downarrow$  & 93.87 & 23.45 & 19.68 & 5.60 & 5.95 & 5.44 & {\ul 5.13} & 5.65 & \textbf{4.92} \\
\midrule
\multirow{4}{*}{\textbf{50}}    & \textbf{PSNR}~(dB)~$\uparrow$  & 14.15 & 27.31 & 28.75 & 39.62 & 39.40 & 39.84 & 40.42 & {\ul 40.52} & \textbf{41.19} \\
                                & \textbf{SSIM}~(\%)~$\uparrow$  & 4.28  & 83.45 & 86.21 & 96.75 & 96.33 & 96.89 & {\ul 97.06} & 96.38 & \textbf{97.73} \\
                                & \textbf{SAM}~(\%)~$\downarrow$ & 96.33 & 24.78 & 21.05 & 6.10 & 7.47 & 5.89 & {\ul 5.51} & 6.57 & \textbf{5.32} \\
                                & \textbf{MAE}~(‰)~$\downarrow$  & 156.45 & 31.20 & 28.15 & 7.15 & 7.56 & 6.93 & {\ul 6.54} & 7.30 & \textbf{6.44} \\
\midrule
\multirow{4}{*}{\textbf{70}}    & \textbf{PSNR}~(dB)~$\uparrow$  & 11.23 & 25.02 & 26.13 & 38.80 & 37.63 & {\ul 39.45} & 38.94 & 38.34 & \textbf{39.68} \\
                                & \textbf{SSIM}~(\%)~$\uparrow$  & 2.34  & 78.12 & 80.53 & 95.80 & 94.72 & {\ul 96.11} & \textbf{96.26} & 94.58 & 96.04 \\
                                & \textbf{SAM}~(\%)~$\downarrow$ & 1.08  & 29.87 & 26.45 & 6.50 & 9.65 & 6.44 & \textbf{6.19} & 9.29 & {\ul 6.32} \\
                                & \textbf{MAE}~(‰)~$\downarrow$  & 2.19  & 38.56 & 35.42 & 9.00 & 9.21 & 8.12 & {\ul 7.73} & 9.29 & \textbf{7.61} \\
\midrule
\multirow{4}{*}{\textbf{30-70}} & \textbf{PSNR}~(dB)~$\uparrow$  & 13.74 & 28.01 & 29.42 & 41.40 & 41.53 & 41.25 & {\ul 42.71} & 42.05 & \textbf{42.87} \\
                                & \textbf{SSIM}~(\%)~$\uparrow$  & 4.62  & 85.60 & 88.35 & 97.25 & 96.90 & 97.48 & {\ul 97.57} & 96.96 & \textbf{97.76} \\
                                & \textbf{SAM}~(\%)~$\downarrow$ & 93.92 & 22.33 & 19.27 & 5.10 & 7.08 & {\ul 5.25} & \textbf{5.02} & 5.45 & 5.66 \\
                                & \textbf{MAE}~(‰)~$\downarrow$  & 155.93 & 29.45 & 26.50 & 6.35 & 6.78 & 5.94 & {\ul 5.67} & 7.63 & \textbf{5.51} \\
                                \midrule
\multirow{4}{*}{\textbf{Realistic}} &\textbf{PSNR}~(dB)~$\uparrow$ & 11.09 & 18.30 & 21.40 & 30.60 & 39.02 & 35.91 & 35.53 & {\ul 39.15} & \textbf{40.31} \\
                                 & \textbf{SSIM}~(\%)~$\uparrow$ & 4.98 & 58.90 & 65.80 & 85.10 & {\ul 97.14} & 96.16 & 96.47 & 96.24 & \textbf{97.83} \\
                                 & \textbf{SAM}~(\%)~$\downarrow$ & 114.35 & 22.10 & 20.90 & 44.20 & 7.84 & 10.96 & 10.35 & {\ul 5.46} & \textbf{5.32} \\
                                 & \textbf{MAE}~(‰)~$\downarrow$ & 189.80 & 30.00 & 28.20 & 51.30 & 7.89 & 10.60 & 10.34 & {\ul 7.75} & \textbf{6.98} \\
\bottomrule
\end{tabular}
\end{table*}

\subsubsection{\textbf{Datasets}}
We conduct denoising experiments on two widely used public datasets: ICVL\footnote{\url{https://icvl.cs.bgu.ac.il/pages/researches/hyperspectral-imaging.html}}~\cite{icvl2016} and Urban\footnote{\url{http://www.escience.cn/people/feiyunZHU/Dataset_GT.html}}~\cite{urbandataset}.
The ICVL dataset was collected by a Specim PS Kappa DX4 hyperspectral camera, consisting of 201 natural scenes with a resolution of $1392\times1300$ pixels across 31 spectral bands (400–700 nm).
The Urban dataset was captured by the HYDICE sensor over an urban scene, comprising $307\times307$ pixels and 162 spectral bands in the 400–2500 nm range~\cite{urbandataset}.

\subsubsection{\textbf{Experimental Settings}}

During training, each ICVL image is divided into non-overlapping patches of size $128\times128\times31$.
For evaluation, each image is centrally cropped to a size of $512\times512$.
We simulate five noise scenarios:
(1) zero-mean Gaussian noise with standard deviation $\sigma=30,50,70$,
(2) non-iid per-band blind Gaussian noise where $\sigma$ is randomly selected from {30, 50, 70} for each spectral band,
and (3) complex realistic noise involving diverse degradation types encountered in practical HSI acquisition.
During training, we apply Gaussian noise randomly sampled from a continuous uniform distribution over the range [10, 70].
Since the Urban dataset does not provide a clean ground truth, we follow the settings in~\cite{NEURIPS2021_2b515e2b} and adopt the APEX dataset~\cite{apexdataset} for pre-training.
In this setup, band-dependent Gaussian noise with standard deviations ranging from 0 to 55 is added to the clean HSIs to simulate realistic degradation.
Our model is trained using the AdamW optimizer with an initial learning rate of $1\mathrm{e}{-4}$ and decays to $1\mathrm{e}{-5}$ after 60 epochs.
The batch size is set to 4, and the training runs for 100 epochs.

\subsubsection{\textbf{Comparative Experiments}}
We compare FairHyp with three model-based methods (BM4D~\cite{bm4d2013}, LRMR~\cite{lrmr2014}, NGMeet~\cite{ng-meet2022}) and four data-driven methods (QRNN3D~\cite{qrnn3d2021}, SST~\cite{sst2023}, SERT~\cite{sert2023}, HSDT~\cite{HSDT2023}).  
As shown in Table~\ref{tab:dn_comp_icvl}, on the ICVL dataset, FairHyp achieves the highest average PSNR, which surpasses the next best method by 0.10 to 0.67 dB and demonstrates superior noise suppression.  
FairHyp also outperforms other methods in terms of SSIM, MAE, and SAM, confirming its ability to preserve both structural and spectral fidelity.
The remarkable and consistent gains confirm that FairHyp provides a robust and generalizable solution for HSI denoising.
Fig.~\ref{fig:dn_vis_icvl} shows a visual comparison on an ICVL sample, with pseudo-color results in upper row and average cumulative error heatmaps in lower row.  
The left focus area emphasizes a patch with low-contrast texture, while the right focus area highlights a smaller region with intricate and fine texture details.
Our method demonstrates the best restoration capabilities.
Fig.~\ref{fig:dn_vis_urban} shows a visual comparison on the Urban dataset with realistic noise.
These results verify that our tri-aspect enhancement strategy effectively balances spatial variability, spectral long-range modeling, and feature efficiency.  

\subsection{HSI Super-resolution} \label{sec:hsis}

\begin{figure*}[t]
    \centering
    \includegraphics[width=\textwidth]{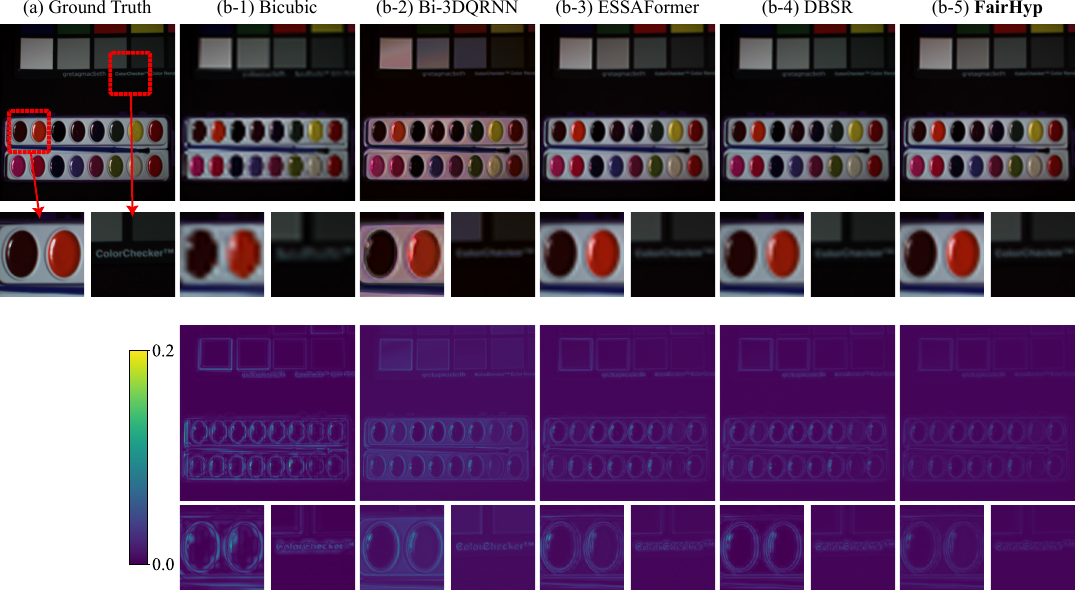}
    \caption{
        \label{fig:sr_vis_cave}
        Comparisons of the super-resolution visual results of different methods on the \textit{CAVE} dataset for \textit{$\times$8}.
    }
\end{figure*}

\begin{table*}[t]
\centering
\caption{Performance comparisons of different methods under super‐resolution scales on the \textit{CAVE} and \textit{Indian Pines} datasets.}
\label{tab:sr_comp}
\begin{tabular}{c|c||r||*{7}{>{\centering\arraybackslash}p{1.3cm}}}
\toprule
\multicolumn{2}{c||}{\textbf{Dataset / Scale}} & \textbf{Metric} 
  & \textbf{Bicubic} & \textbf{SSPSR} & \textbf{MCNet} & \textbf{Bi3DQRNN} 
  & \textbf{ESSAFormer} & \textbf{DBSR} & \textbf{FairHyp} \\
\midrule
\multirow{8}{*}{\textbf{CAVE}}
  & \multirow{4}{*}{$\times4$} 
    & PSNR~(dB)~$\uparrow$    & 34.83 & 35.89 & 37.48 & 37.66 & 39.21 & {\ul 39.27} & \textbf{40.43} \\
  &                              
    & SSIM~(\%)~$\uparrow$     & 92.84 & 93.63 & 94.86 & 93.72 & {\ul 95.58} & 95.51 & \textbf{96.64} \\
  &                              
    & SAM~(\%)~$\downarrow$    & 8.79  & 9.46  & 9.21  & {\ul 8.66}  & 9.08  & 8.67  & \textbf{8.37}  \\
  &                              
    & MAE~(‰)~$\downarrow$     & 9.42  & 8.43  & 5.88  & 7.54  & 6.87  & {\ul 5.41}  & \textbf{4.22}  \\
\cmidrule{2-10}
  & \multirow{4}{*}{$\times8$} 
    & PSNR~(dB)~$\uparrow$    & 30.36 & 32.73 & 32.18 & 32.84 & 33.87 & {\ul 34.19} & \textbf{35.11} \\
  &                              
    & SSIM~(\%)~$\uparrow$     & 84.97 & 85.93 & 86.62 & 86.68 & 87.91 & {\ul 89.28} & \textbf{90.88} \\
  &                              
    & SAM~(\%)~$\downarrow$    & 22.87 & 17.38 & 17.45 & 16.78 & 14.52 & {\ul 11.02} & \textbf{9.81}  \\
  &                              
    & MAE~(‰)~$\downarrow$     & 16.42 & 12.01 & 11.23 & 9.72  & {\ul 9.64}  & 10.16 & \textbf{9.48}  \\
\midrule
\midrule
\multirow{8}{*}{\textbf{Chikusei}}
  & \multirow{4}{*}{$\times4$} 
    & PSNR~(dB)~$\uparrow$    & 34.10 & 35.17 & 37.24 & 38.38 & 39.43 & {\ul 39.56} & \textbf{41.79} \\
  &                              
    & SSIM~(\%)~$\uparrow$     & 89.08 & 89.86 & 90.54 & 90.11 & 91.23 & {\ul 92.39} & \textbf{93.47} \\
  &                              
    & SAM~(\%)~$\downarrow$    & 4.33  & 5.07  & 3.86  & 4.41  & 3.76  & {\ul 3.32}  & \textbf{3.19}  \\
  &                              
    & MAE~(‰)~$\downarrow$     & 10.59 & 9.53  & {\ul 6.27}  & 8.36  & 7.43  & 7.14  & \textbf{5.08}  \\
\cmidrule{2-10}
  & \multirow{4}{*}{$\times8$} 
    & PSNR~(dB)~$\uparrow$    & 31.67 & 33.15 & 32.83 & 33.46 & 34.28 & {\ul 34.59} & \textbf{35.38} \\
  &                              
    & SSIM~(\%)~$\uparrow$     & 87.08 & 88.27 & 88.65 & 88.76 & {\ul 90.53} & 89.78 & \textbf{91.86} \\
  &                              
    & SAM~(\%)~$\downarrow$    & 5.59  & 4.67  & 4.93  & 4.53  & 4.26  & {\ul 3.82}  & \textbf{3.57}  \\
  &                              
    & MAE~(‰)~$\downarrow$     & 14.31 & 12.25 & 11.57 & 10.27 & 10.16 & {\ul 9.26}  & \textbf{8.09}  \\
\bottomrule
\end{tabular}
\end{table*}

\subsubsection{\textbf{Datasets}} \label{hsis}

We conduct super-resolution experiments on two widely used public datasets: CAVE\footnote{\url{http://www.cs.columbia.edu/CAVE/databases/}}~\cite{cave} and Chikusei\footnote{\url{https://naotoyokoya.com/Download.html}}~\cite{chikusei}.
The CAVE dataset consists of 32 natural hyperspectral images with spatial dimensions of $512 \times 512$ and 31 spectral bands, covering wavelengths from 400 nm to 700 nm, acquired using a generalized assorted pixel camera.
The Chikusei dataset was captured by the Headwall Hyperspec-VNIR-C imaging sensor over agricultural and urban regions in Chikusei, Ibaraki, Japan, on July 29, 2014, between 9:56 and 10:53 UTC+9.
The central point of the scene is located at coordinates 36.294946N, 140.008380E.
It contains 128 spectral bands spanning the range from 363 nm to 1018 nm.
The scene comprises $2517 \times 2335$ pixels, with a ground sampling distance of 2.5 meters.

\subsubsection{\textbf{Experimental Settings}}

Each CAVE image is divided into non-overlapping cubes of size $128\times128$ during training.
For the Chikusei dataset, we extract 20 non-overlapping patches of size $128 \times 128$ that contain valid image content.  
Among these, 10 patches are randomly selected as the training set, while the remaining 10 are used for testing.
We conduct super-resolution experiments with two upscaling factors: $4\times$ and $8\times$.  
The low-resolution inputs are generated by bicubic downsampling the high-resolution images along the spatial dimensions.  
All models, including FairHyp, are trained using the AdamW optimizer.  
The learning rate is initialized at $1\mathrm{e}{-4}$ and decays to $1\mathrm{e}{-5}$ after 60 epochs.  
The batch size is set to 4, and training is conducted for 100 epochs.  

\subsubsection{\textbf{Comparative Experiments}}

We compare our FairHyp with other five methods including SSPSR~\cite{sspn2020}, MCNet~\cite{mcnet2020}, Bi-3DQRNN~\cite{bi3dqrnn2021}, ESSAFormer~\cite{essaformer2023}, and DBSR~\cite{dbsr2025}.
Besides, we employ the bicubic interpolation as a baseline method.
Table~\ref{tab:sr_comp} presents the quantitative comparison results under different super-resolution scales on the \textit{CAVE} and \textit{Indian Pines} datasets.
FairHyp consistently delivers the highest performance across all evaluated metrics and scales.
On the \textit{CAVE} dataset at $\times4$ scale, it achieves a PSNR of 40.43~dB, surpassing the second-best method, DBSR, by a margin of 1.16~dB.
Similarly, its SSIM score of 96.64\% reflects a notable improvement in structural preservation compared to ESSAFormer’s 95.58\%.
For spectral accuracy, FairHyp obtains the lowest SAM (8.37\%) and MAE (4.22‰), indicating superior spectral fidelity and reduced absolute error.
This performance advantage is also evident at $\times8$ scale, where FairHyp achieves 35.11~dB in PSNR and 90.88\% in SSIM, significantly outperforming other advanced methods such as DBSR and ESSAFormer.
On the \textit{Indian Pines} dataset, FairHyp maintains its leading position.
At $\times4$ scale, it records 41.79~dB in PSNR and 93.47\% in SSIM, reflecting high reconstruction quality.
Its SAM and MAE values, 3.19\% and 5.08‰ respectively, further confirm its effectiveness in preserving spectral and pixel-level accuracy.
Even under the more challenging $\times8$ scale, FairHyp achieves the best performance across all metrics, highlighting its robustness and generalizability.
Overall, these results demonstrate FairHyp’s strong capability in HSI super-resolution tasks, achieving both high-fidelity reconstructions and accurate spectral recovery across diverse datasets and scaling factors.
Fig.~\ref{fig:sr_vis_cave} presents $\times 8$ super-resolution results on the CAVE dataset.
Subfigures (b)–(f) show pseudo-RGB reconstructions, while subfigures (g)–(k) display cumulative error maps over all spectral bands.
FairHyp demonstrates superior detail preservation compared to other methods.

\subsection{HSI Inpainting}

\begin{table*}[t]
    \centering
    \caption{Quantitative comparison of inpainting results on the \textit{ICVL} and \textit{Chikusei} datasets.}
    \label{tab:inp_comp}
    \begin{tabular}{c|*{1}{>{\raggedleft\arraybackslash}p{1.8cm}}||*{1}{>{\centering\arraybackslash}p{1.1cm}}|*{3}{>{\centering\arraybackslash}p{1.1cm}}|*{4}{>{\centering\arraybackslash}p{1.1cm}}|*{1}{>{\centering\arraybackslash}p{1.1cm}}}
        \toprule
        \multicolumn{3}{c|}{\textbf{}} & \multicolumn{3}{c|}{\textbf{Model Driven}} & \multicolumn{4}{c|}{\textbf{Data Driven}} & \textbf{Ours} \\
        \midrule
        \textbf{Dataset} & Metrics & Corrupted & FastHyIn & DHSP2D & WLRTR & QRNN3D & SST & SERT & HSDT & \textbf{FairHyp} \\
        \midrule
        \multirow{4}{*}{\textbf{ICVL}} & \textbf{PSNR}~(dB)~$\uparrow$    & 26.77 & 35.50 & 36.80 & 38.20 & 42.58 & 41.31 & 42.94 & {\ul 43.11} & \textbf{43.53} \\
        & \textbf{SSIM}~(\%)~$\uparrow$                                   & 95.43 & 95.80 & 96.30 & 96.90 & {\ul 98.59} & 98.54 & 98.28 & 98.19 & \textbf{99.16} \\
        & \textbf{SAM}~(\%)~$\downarrow$                                   & 7.34 & 5.50 & 4.90 & 4.30 & 2.36 & {\ul 2.11} & 2.27 & 2.28 & \textbf{1.84} \\
        & \textbf{MAE}~(‰)~$\downarrow$                                   & 5.51 & 6.90 & 6.30 & 5.80 & 5.65 & 5.72 & {\ul 5.06} & 5.14 & \textbf{4.94} \\
        \midrule
        \midrule
        \multirow{4}{*}{\textbf{Chikusei}} & \textbf{PSNR}~(dB)~$\uparrow$   & 25.99 & 36.20 & 37.10 & 38.10 & 39.87 & 39.87 & {\ul 40.31} & 39.85 & \textbf{40.92} \\
        & \textbf{SSIM}~(\%)~$\uparrow$                                      & 91.09 & 91.20 & 91.65 & 92.00 & 92.15 & 91.27 & 94.19 & {\ul 95.08} & \textbf{96.05} \\
        & \textbf{SAM}~(\%)~$\downarrow$                                      & 28.37 & 24.00 & 21.70 & 20.10 & 20.06 & 31.08 & 26.96 & {\ul 14.77} & \textbf{14.68} \\
        & \textbf{MAE}~(‰)~$\downarrow$                                      & 4.12 & 9.50 & 8.60 & 7.80 & 7.03 & 7.13 & {\ul 5.96} & 6.12 & \textbf{5.27} \\
        \bottomrule
    \end{tabular}
\end{table*}

\begin{figure*}[t]
    \centering
    \includegraphics[width=\textwidth]{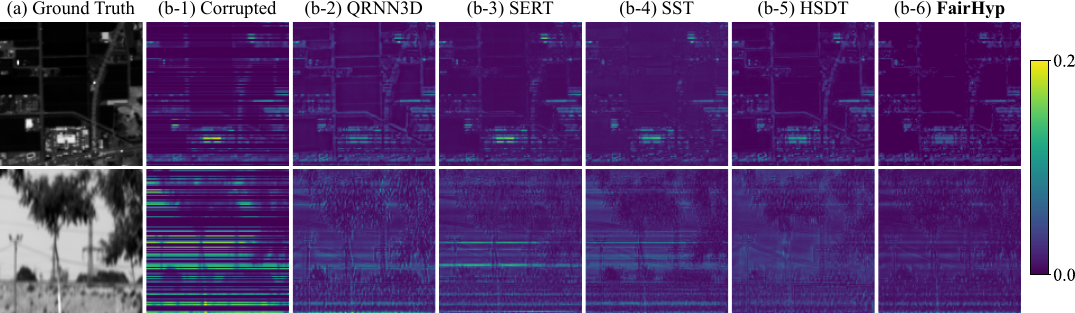}
    \caption{
        \label{fig:inp_vis}
        Comparisons of the inpainting visual results of different methods on the \textit{Chikusei} dataset (up) and \textit{CAVE} dataset (below).}
\end{figure*}

\subsubsection{\textbf{Datasets}}
We conduct inpainting experiments on two widely used public datasets: Chikusei and ICVL, which has already been introduced in details in Section~\ref{sec:hsic} and Section~\ref{sec:hsid}, repectively.

\subsubsection{\textbf{Experimental Settings}}

During training, each ICVL image is divided into non-overlapping cubes of size $128\times128$.
For evaluation, each image is centrally cropped to a size of $512\times512$.
For the Chikusei dataset, we extract 20 non-overlapping patches of size $128 \times 128$ that contain valid image content.  
Among these, 10 patches are randomly selected as the training set, while the remaining 10 are used for testing.
To simulate the incompleteness problem of HSI, we randomly select one-third of the bands in each HSI and introduce deadline noise, affecting 10\% to 30\% of the rows in each selected band.
All model are trained using the AdamW optimizer with an initial learning rate of $1\mathrm{e}{-4}$.
The learning rate is initialized at $1\mathrm{e}{-4}$ and decays to $1\mathrm{e}{-5}$ after 60 epochs.
The batch size is set to 4, and the training runs for 100 epochs.

\subsubsection{\textbf{Comparative Experiments}}

We evaluate our method in the HSI inpainting task against three model-based methods and seven deep learning–based methods.  
We compare FairHyp with three model-based methods (FastHyIn~\cite{fasthyin2018}, Deep HS Preior 2D~\cite{dhp2d2019} (DHSP2D), WLRTR~\cite{wlrtr2020}) and four data-driven methods (QRNN3D~\cite{qrnn3d2021}, SST~\cite{sst2023}, SERT~\cite{sert2023}, HSDT~\cite{HSDT2023}).
As shown in Table~\ref{tab:inp_comp}, FairHyp consistently outperforms both model-driven and data-driven approaches across all evaluation metrics on the ICVL and Chikusei datasets.
In particular, FairHyp achieves the highest PSNR and SSIM values, indicating superior reconstruction fidelity and structural preservation.
Furthermore, it obtains the lowest SAM and MAE scores, demonstrating its capability to preserve spectral consistency and reduce pixel-wise errors.
These improvements are especially prominent in the Chikusei dataset, where FairHyp shows a significant advantage in handling complex outdoor scenes.
Visual results in Fig.~\ref{fig:inp_vis} further confirm that FairHyp yields cleaner reconstructions with fewer artifacts and better spectral integrity compared to other methods.
This superior performance is attributed to the synergistic design of FairHyp’s tri-aspect modules, which jointly address spatial variability, spectral irregularity, and feature sparsity.
Overall, the results validate the robustness and generalizability of FairHyp in restoring missing information across diverse hyperspectral imaging conditions.

\subsection{Further Analysis and Explorations}

To further explore the internal mechanism and practical advantages, we conduct more experiments.
This section provides targeted evaluations on effectiveness, efficiency, and the core capabilities of each module, offering deeper insights into how FairHyp addresses the challenges of non-uniformity in hyperspectral representation.

\subsubsection{\textbf{Evaluation of Overall Effectiveness}}

\begin{table*}[ht]
    \centering
    \caption{Module-wise ablation study of FairHyp across four HSI tasks.. \label{tab:overallablation}}
    \begin{tabular}{*{3}{>{\centering\arraybackslash}p{1.2cm}}||*{3}{>{\centering\arraybackslash}p{1.2cm}}||*{4}{>{\centering\arraybackslash}p{1.2cm}}}
        \toprule
        \multicolumn{3}{c||}{\textbf{Settings}} & \multicolumn{3}{c||}{\textbf{Classification (on \textit{Indian Pines})}} & \multicolumn{4}{c}{\textbf{Denoising (on \textit{ICVL} with \textit{Blind Gaussian} noise)}} \\
        \midrule
        \multirow{2}{*}{\textit{RK4-SVA}} & \multirow{2}{*}{\textit{S$^2$FairConv}} & \multirow{2}{*}{\textit{SCSS}} & \textbf{OA} & \textbf{AA} & \textbf{$\kappa$} & \textbf{PSNR} & \textbf{SSIM} & \textbf{SAM} & \textbf{MAE} \\
         & & & (\%) $\uparrow$ & (\%) $\uparrow$ & ($\times 10^2$) $\uparrow$ & (\%) $\uparrow$ & (\%) $\uparrow$ & (\%) $\downarrow$ & (‰) $\downarrow$\\
        \midrule
        $\times$ & $\times$ & $\times$             & 89.22 & 84.26 & 87.68 & 39.75 & 93.88 & 7.91 & 7.45 \\
        $\checkmark$ & $\times$ & $\times$         & 92.32 & 89.20 & 91.04 & 41.12 & 96.25 & 6.90 & 6.49 \\
        $\times$ & $\checkmark$ & $\times$         & 93.33 & 92.01 & 92.67 & 42.08 & 97.05 & 5.78 & 5.94 \\
        $\times$ & $\times$ & $\checkmark$         & 92.72 & 93.81 & 92.02 & 41.90 & 96.71 & 6.02 & 6.12 \\
        $\checkmark$ & $\times$ & $\checkmark$     & 96.15 & 95.38 & 95.92 & 42.15 & 97.20 & 5.81 & 5.79 \\
        $\checkmark$ & $\checkmark$ & $\checkmark$ & \textbf{97.76} & \textbf{96.47} & \textbf{97.53} & \textbf{42.87} & \textbf{97.76} & \textbf{5.66} & \textbf{5.51} \\
        \bottomrule     
    \end{tabular}

    \vspace{0.15cm}
    \begin{tabular}{*{3}{>{\centering\arraybackslash}p{1.2cm}}||*{4}{>{\centering\arraybackslash}p{1.1cm}}||*{4}{>{\centering\arraybackslash}p{1.1cm}}}
        \toprule
        \multicolumn{3}{c||}{\textbf{Settings}} & \multicolumn{4}{c}{\textbf{Super-Resolution (on $\times 4$ \textit{CAVE})}} & \multicolumn{4}{c}{\textbf{Inpainting (on \textit{Chikusei})}} \\
        \midrule
        \multirow{2}{*}{\textit{RK4-SVA}} & \multirow{2}{*}{\textit{S$^2$FairConv}} & \multirow{2}{*}{\textit{SCSS}} & \textbf{PSNR} & \textbf{SSIM} & \textbf{SAM} & \textbf{MAE}& \textbf{PSNR} & \textbf{SSIM} & \textbf{SAM} & \textbf{MAE} \\
         & & & (\%) $\uparrow$ & (\%) $\uparrow$ & (\%) $\downarrow$ & (‰) $\downarrow$ & (\%) $\uparrow$ & (\%) $\uparrow$ & (\%) $\downarrow$ & (‰) $\downarrow$\\
        \midrule
        $\times$ & $\times$ & $\times$             & 37.14 & 92.90 & 10.72 & 5.84 & 37.97 & 92.44 & 19.26 & 6.87 \\
        $\checkmark$ & $\times$ & $\times$         & 38.70 & 95.33 & 9.94  & 5.31 & 39.50 & 94.21 & 16.92 & 6.13 \\
        $\times$ & $\checkmark$ & $\times$         & 39.72 & 96.08 & 8.82  & 4.89 & 40.18 & 95.21 & 15.98 & 5.71 \\
        $\times$ & $\times$ & $\checkmark$         & 39.01 & 95.88 & 9.35  & 5.20 & 39.41 & 94.65 & 17.12 & 6.10 \\
        $\checkmark$ & $\times$ & $\checkmark$     & 40.15 & 96.32 & 8.75  & 4.58 & 40.45 & 95.90 & 15.31 & 5.41 \\
        $\checkmark$ & $\checkmark$ & $\checkmark$ & \textbf{40.43} & \textbf{96.64} & \textbf{8.37} & \textbf{4.22} & \textbf{40.92} & \textbf{96.05} & \textbf{14.68} & \textbf{5.27} \\
        \bottomrule     
    \end{tabular}
\end{table*}

To evaluate the contribution of each module in FairHyp, we perform an ablation study by selectively enabling RK4-SVA, S$^2$FairConv, and SCSS across four representative hyperspectral tasks.
As shown in Table~\ref{tab:overallablation}, the removal of any individual component results in a consistent performance degradation, confirming the functional importance of each design.
The SCSS module plays a central role in spectral modeling, with clear improvements observed in denoising and super-resolution.
RK4-SVA improves spatial continuity and classification accuracy, particularly under spatially irregular conditions such as inpainting.
S$^2$FairConv enhances feature representation quality, contributing across all tasks.
The full configuration combining all modules yields the best results, highlighting the complementary nature of spatial adaptation, spectral modeling, and feature integration.
These results demonstrate the proposed tri-aspect design effectively addresses the structural non-uniformity in hyperspectral data, which is often overlooked by uniform architectures that apply equal treatment across dimensions.

\subsubsection{\textbf{Analysis of Computational Efficiency}}

\begin{table}[th]
    \caption{Cost comparison with other state-of-the-art restoration methods.} \label{tab:eff1}
    \centering
    \begin{tabular}{l||*{3}{>{\centering\arraybackslash}p{1.2cm}}}
    \toprule
    \multirow{2}{*}{\textbf{Method}} & \textbf{FLOPs} & \textbf{Paras.} & \textbf{Time} \\
    & (G)~$\downarrow$ & (M)~$\downarrow$ & (ms)~$\downarrow$\\
    \midrule
    \midrule
    \multicolumn{4}{c}{\textit{Denoising \& Inpainting} ($128 \times 128$ with 31 bands)} \\
    \midrule
    SST       & 75.43 & 4.14 & 144.70 \\
    SERT      & \textbf{33.88}  & 1.90 & 44.46 \\
    QRNN3D    & 78.66 & 0.86 & 34.72 \\
    HSDT      & 46.98 & \textbf{0.52} & 48.90 \\
    \midrule
    FairHyp   & {\ul 37.41}  & {\ul 0.64} & \textbf{25.62} \\
    \midrule
    \midrule
    \multicolumn{4}{c}{\textit{Super-Resolution} ($64$ to $512$ with 31 bands)} \\
    \midrule
    MCNet     & {\ul 278.34} & 2.17 & {\ul 118.87} \\
    SSPSR     & 749.13 & 26.08 & 127.92 \\
    ESSAFormer& \textbf{205.63} & {\ul 0.88} & \textbf{96.11} \\
    DBSR      & 751.60 & 2.52 & 226.04 \\
    \midrule
    FairHyp   & 598.51  & \textbf{0.81} & 122.76 \\
    \bottomrule
    \end{tabular} 
\end{table}

\begin{table}[th]
    \caption{Cost comparison with other state-of-the-art classification methods.} \label{tab:eff2}
    \centering
    \begin{tabular}{l||*{3}{>{\centering\arraybackslash}p{1.2cm}}}
    \toprule
    \multirow{2}{*}{\textbf{Method}} & \textbf{FLOPs} & \textbf{Paras.} & \textbf{Time} \\
    & (M)~$\downarrow$ & (M)~$\downarrow$ & (ms)~$\downarrow$\\
    \midrule
    \midrule
    \multicolumn{4}{c}{\textit{Classification} (One patch with 200 bands)} \\
    \midrule
    CSCANet             & 691.68 & 7.37 & 1.88 \\
    SSFTT               & 47.76 & 0.95 & \textbf{2.44} \\
    SpectralFormer      & 61.39 & {\ul 0.36} & 2.58 \\
    DSFormer            & {\ul 43.73} & 0.87 & 7.35 \\
    \midrule
    FairHyp             & 54.07  & 0.82 & 3.44 \\
    FairHyp*            & \textbf{15.58}  & \textbf{0.33} & {\ul 2.51} \\
    \bottomrule
    \end{tabular} 
\end{table}

\begin{figure}[t]
    \centering
    \begin{subfigure}{0.45\textwidth}
        \centering
        \begin{subfigure}[b]{1\textwidth}
            \caption{OA vs. Cost on Indian Pines.}
            \centering
            \includegraphics[width=\textwidth]{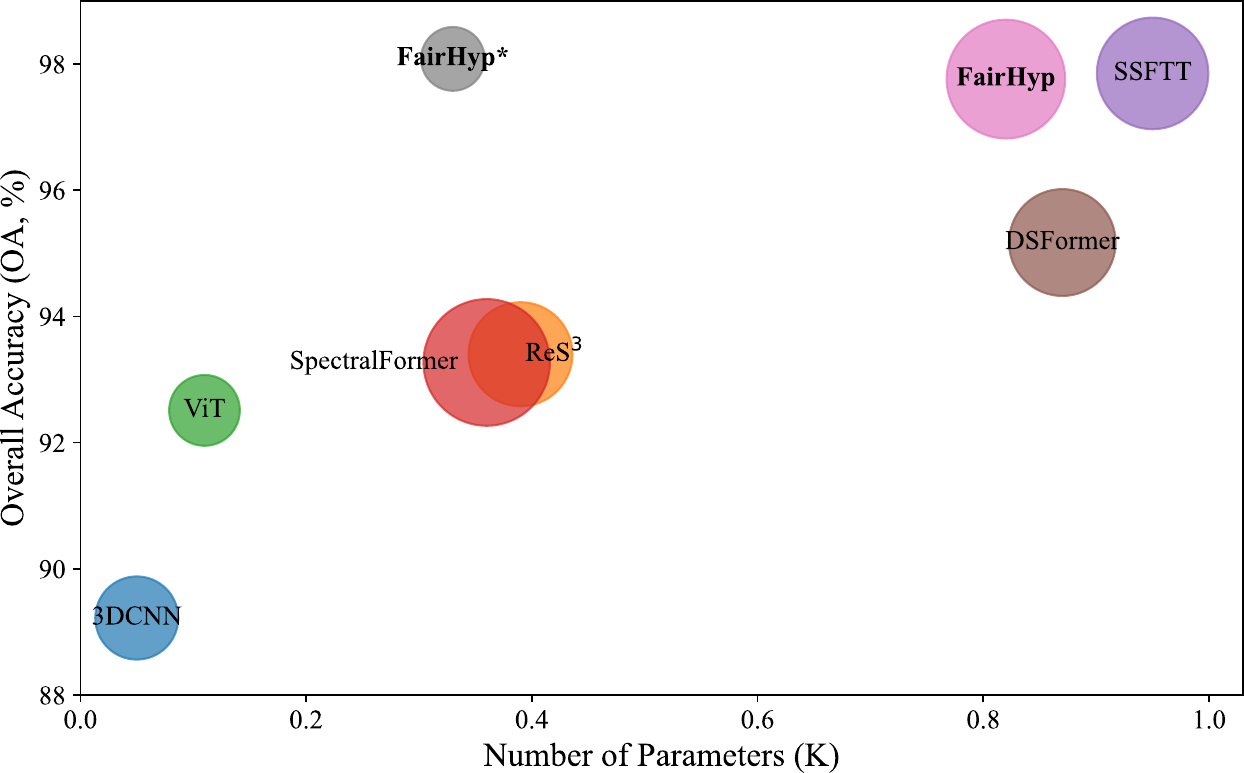}
        \end{subfigure}
        \hfill
        \begin{subfigure}[b]{1\textwidth}
            \caption{AA vs. Cost on WHU-Hi-HongHu.}
            \centering
            \includegraphics[width=\textwidth]{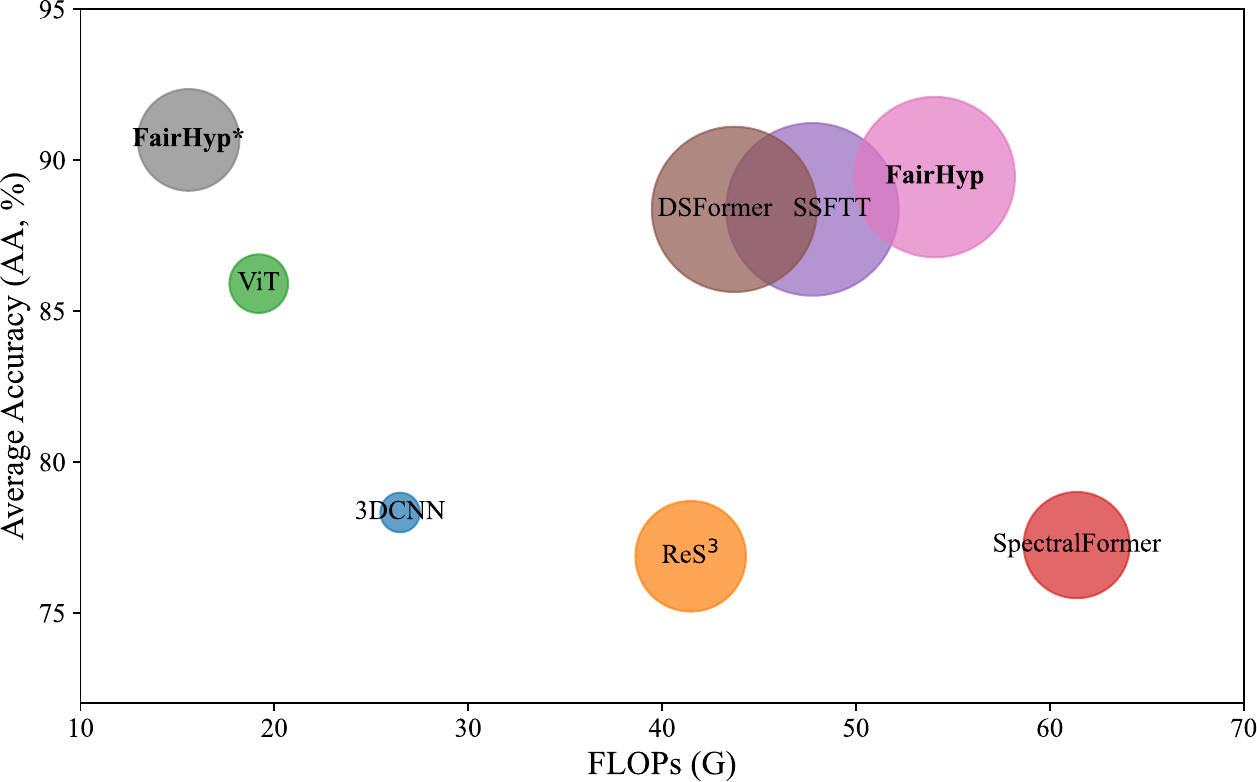}
        \end{subfigure}
    \end{subfigure}

    \caption{
        Comparison of classification performance, model size, and computational complexity.
        The x-axis denotes the number of parameters, the y-axis represents performence, and the circle area indicates FLOPs.
    }
    \label{fig-cls_efficiency}
\end{figure}

As demonstrated in previous sections, FairHyp consistently achieves state-of-the-art performance across restoration and classification tasks.
Unlike task-specific designs, FairHyp serves as a general-purpose framework for various hyperspectral restoration scenarios.
As shown in Tables~\ref{tab:eff1} and~\ref{tab:eff2}, although it does not always yield the lowest FLOPs or runtime, its overall efficiency remains competitive.
In particular, it achieves the fastest inference in denoising, competitive cost in super-resolution, and the smallest model size in classification (FairHyp*).
To further illustrate its advantage in classification, Fig.~\ref{fig-cls_efficiency} visualize the trade-off between performance and computational cost.
FairHyp and FairHyp* achieve superior accuracy while maintaining lower complexity, with FairHyp* especially excelling in both parameter count and FLOPs.
This highlights the method's practical value for deployment in resource-constrained settings.


\subsubsection{\textbf{Multi-Receptive Field Capability of \textit{S$^2$FairConv}}}

\begin{table*}[ht]
    \centering
    \caption{Effect of different receptive field in \textit{S$^2$FairConv}. \label{tab:multireceptivefield}}
    \begin{tabular}{*{1}{>{\centering\arraybackslash}p{1.2cm}}|*{1}{>{\centering\arraybackslash}p{1.2cm}}|*{1}{>{\centering\arraybackslash}p{1.2cm}}||*{3}{>{\centering\arraybackslash}p{1.2cm}}||*{4}{>{\centering\arraybackslash}p{1.2cm}}}
        \toprule
        \multicolumn{3}{c||}{\textbf{Settings}} & \multicolumn{3}{c||}{\textbf{Classification (on \textit{Indian Pines})}} & \multicolumn{4}{c}{\textbf{Super-Resolution (on $\times 4$ \textit{CAVE})}} \\
        \midrule
        \multicolumn{2}{c|}{\textbf{Local}} & \textbf{Global} & \textbf{OA} & \textbf{AA} & \textbf{$\kappa$} & \textbf{PSNR} & \textbf{SSIM} & \textbf{SAM} & \textbf{MAE} \\
        \textbf{Spatial} & \multicolumn{2}{c||}{\textbf{Spectral}} & (\%) $\uparrow$ & (\%) $\uparrow$ & ($\times 10^2$) $\uparrow$ & (\%) $\uparrow$ & (\%) $\uparrow$ & (\%) $\downarrow$ & (‰) $\downarrow$\\
        \midrule
        $\times$ & $\times$ & $\times$ & 93.55 & 91.22 & 92.10 & 37.48 & 94.88 & 12.54 & 6.21 \\
        $\times$ & $\checkmark$ & $\times$ & 94.93 & 93.44 & 94.10 & 38.77 & 95.60 & 10.63 & 5.77 \\
        $\checkmark$ & $\times$ & $\times$ & 96.12 & 94.90 & 95.58 & 38.32 & 95.37 & 10.91 & 5.84 \\
        $\times$ & $\times$ & $\checkmark$ & 94.15 & 93.01 & 93.35 & 39.91 & 96.08 & 9.41  & 5.08 \\
        $\checkmark$ & $\times$ & $\checkmark$ & 96.48 & 95.02 & 95.91 & 40.05 & 96.33 & 9.01  & 4.79 \\
        $\checkmark$ & $\checkmark$ & $\times$ & 97.08 & 96.01 & 96.72 & 39.22 & 96.20 & 9.63  & 4.99 \\
        $\times$ & $\checkmark$ & $\checkmark$ & 95.11 & 94.00 & 94.35 & 40.21 & 96.49 & 8.75  & 4.44 \\
        $\checkmark$ & $\checkmark$ & $\checkmark$ & \textbf{97.76} & \textbf{96.47} & \textbf{97.53} & \textbf{40.43} & \textbf{96.64} & \textbf{8.37} & \textbf{4.22} \\
        \bottomrule       
    \end{tabular}
\end{table*}

To evaluate the effectiveness of multi-receptive field design in \textit{S$^2$FairConv}, we conduct ablation experiments by selectively enabling local spatial, local spectral, and global spectral receptive fields.
As shown in Table~\ref{tab:multireceptivefield}, enabling any individual component leads to performance gains over the baseline with no receptive field modeling, indicating the distinct contributions of each design.
Specifically, incorporating global spectral modeling significantly improves PSNR and SAM, validating its advantage in capturing long-range spectral dependencies for super-resolution tasks.
Local spatial modeling, on the other hand, leads to notable improvements in classification accuracy (OA = 96.12\%), highlighting the importance of spatial context for semantic discrimination.
When both local and global receptive fields are combined, further performance gains are observed across both tasks, reflecting their complementary roles.
The full configuration, integrating all three receptive field types, achieves the best results in both classification (OA = 97.76\%) and reconstruction (PSNR = 40.43), confirming that \textit{S$^2$FairConv} benefits from a synergistic multi-scale spatial-spectral design.
This demonstrates that various and more globally spectral receptive field is crucial for modeling the information of hyperspectral data.

\subsubsection{\textbf{Feature Efficiency Enhancement via \textit{S$^2$FairConv}}}

\begin{table}[th]
    \centering
    \caption{Effect of feature skipping in \textit{S$^2$FairConv} under varying feature dimensions.}
    \label{tab:featureefficiency}
    \begin{tabular}{l|c||cc||cccc}
        \toprule
        \multicolumn{2}{c||}{\textbf{Settings}} &
          \textbf{FLOPs} &
          \textbf{Paras.} &
          \textbf{PSNR} &
          \textbf{SSIM} &
          \textbf{SAM} &
          \textbf{MAE} \\
        \textbf{FS} & \textbf{Dim.} &
          (G) $\downarrow$ &
          (M) $\downarrow$ &
          (dB) $\uparrow$ &
          (\%) $\uparrow$ &
          (\%) $\downarrow$ &
          (‰) $\downarrow$ \\
        \midrule
        \multirow{4}{*}{\rotatebox{90}{\textit{w/o} \textbf{FS}}} 
            & 4  & 11.22 & 0.31 & 41.25 & 93.92 & 22.33 & 29.45 \\
            & 8  & 42.42 & 0.73 & 42.48 & 97.20 & 5.45 & 5.95 \\
            & 12 & \cellcolor{gray!15}82.45 & \cellcolor{gray!15}1.57
              & \cellcolor{gray!15}\textbf{42.66} & \cellcolor{gray!15}\textbf{97.57}
              & \cellcolor{gray!15}\textbf{5.30} & \cellcolor{gray!15}\textbf{5.62} \\
            & 16 & 146.31 & 2.89 & 42.59 & 97.31 & 5.48 & 5.80 \\
        \midrule
        \multirow{4}{*}{\rotatebox{90}{\textit{with} \textbf{FS}}} 
            & 4  & 9.97 & 0.28 & 41.80 & 96.90 & 7.08 & 6.78 \\
            & 8  & \cellcolor{gray!15}37.41 & \cellcolor{gray!15}0.64
              & \cellcolor{gray!15}\textbf{42.76} & \cellcolor{gray!15}\textbf{97.65}
              & \cellcolor{gray!15}\textbf{5.10} & \cellcolor{gray!15}\textbf{5.49} \\
            & 12 & 79.54 & 1.48 & 42.64 & 97.44 & 5.22 & 5.61 \\
            & 16 & 138.24 & 2.40 & 42.66 & 97.50 & 5.25 & 5.58 \\
        \bottomrule     
    \end{tabular}
\end{table}

To assess the feature efficiency of the proposed \textit{S$^2$FairConv}, we conduct a study by comparing models with and without feature skipping (FS) connections under varying channel dimensions.
As shown in Table~\ref{tab:featureefficiency}, the model equipped with FS at a low dimension of 8 achieves the highest overall performance across all metrics, including the best PSNR (42.76~dB), SSIM (97.65\%), SAM (5.10\%), and MAE (5.49‰), while maintaining the lowest computational cost among high-performing configurations (37.41 GFLOPs, 0.64 M parameters).
Notably, in the absence of FS, the optimal performance is observed at a higher dimension of 12 (PSNR = 42.66~dB), which demands more than 82.45 GFLOPs and 1.57 M parameters.
This contrast highlights that \textit{S$^2$FairConv} with FS can achieve superior reconstruction quality using fewer features and lower computational overhead.
These results validate that the feature skipping mechanism enhances the efficiency of feature utilization, enabling the model to maintain or even improve performance under constrained dimensionality.
Such improvements confirm the role of \textit{S$^2$FairConv} in facilitating compact and expressive hyperspectral representations.

\subsubsection{\textbf{Global Spectral Context Modeling with \textit{SCSS}}}

\begin{figure*}[t]
    \centering
    \includegraphics[width=\textwidth]{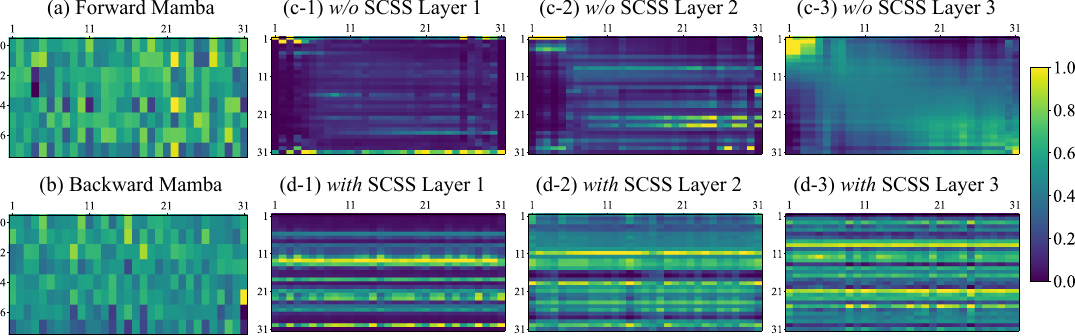}
    \caption{
        \label{fig:fig_scss_mid}
        Visualization of spectral context modeling in \textit{SCSS}.
        Attention patterns from Forward/Backward Mamba and ablation of SCSS layers on the \textit{ICVL} dataset.
        SCSS enables structured and coherent spectral dependencies across layers.
    }
\end{figure*}

\begin{table}[ht]
    \centering
    \caption{Effect of different spectral context settings of \textit{SCSS}.
    Ablation on static statistics (Sta.), forward (F.W.), backward (B.W.), and bidirectional (Bi.D.) SCSS reveals their impact on spectral context enhancement.}
    \label{tab:globalcontext}
    \begin{tabular}{c|c||cc||cccc}
        \toprule
        \multicolumn{2}{c||}{\multirow{2}{*}{\textbf{Settings}}} &
          \textbf{FLOPs} &
          \textbf{Paras.} &
          \textbf{PSNR} &
          \textbf{SSIM} &
          \textbf{SAM} &
          \textbf{MAE} \\
        \multicolumn{2}{c||}{}&
          (G) $\downarrow$ &
          (M) $\downarrow$ &
          (dB) $\uparrow$ &
          (\%) $\uparrow$ &
          (\%) $\downarrow$ &
          (‰) $\downarrow$ \\
        \midrule
            \multicolumn{2}{c||}{Pure CNN}             & 32.17 & 0.59 & 40.79 & 95.58 & 11.92 & 10.87\\
            \multicolumn{2}{c||}{\textit{w/o} SCSS}    & 36.81 & 0.63 & 42.18 & 97.17 & 6.02 & 5.68 \\
            \midrule
            \multirow{4}{*}{\rotatebox{90}{\textit{with} SCSS}} & \textit{Sta.} & 37.02 & 0.64 & 42.35 & 97.28 & 5.85 & 5.62 \\
                                                                 & \textit{F.W.} & \multirow{2}{*}{37.21} & \multirow{2}{*}{0.64} & 42.55 & 97.40 & 5.72 & 5.56 \\
                                                                 & \textit{B.W.} &  &  & 42.57 & 97.42 & 5.70 & 5.54 \\
             & \textit{Bi.D.} & 37.41 & 0.64 & \textbf{42.87} & \textbf{97.76} & \textbf{5.66} & \textbf{5.51} \\
        \bottomrule     
    \end{tabular}
\end{table}

To better understand the role of SCSS in spectral modeling, we visualize intermediate attention responses, as shown in Fig.~\ref{fig:fig_scss_mid}.
Subfigures (a) and (b) depict the response maps produced by forward and backward Mamba scanning, respectively.
The two exhibit distinctly different activation distributions across spectral bands, with minimal overlap in response patterns.
This indicates that forward and backward scanning are not redundant.
Instead, they capture asymmetric spectral dependencies in opposite directions, which justifies their coexistence in the full SCSS design.
Subfigures (c-1) to (c-3) correspond to a simplified channel-wise self-attention mechanism.
The resulting activations are largely concentrated around diagonal regions, suggesting that the model predominantly captures local band relationships.
In contrast, subfigures (d-1) to (d-3), representing the complete SCSS configuration, exhibit clear spectral coherence along the vertical axis.
Several non-contiguous band regions exhibit similar or mirrored activation patterns, indicating that the model has captured long-range and potentially periodic dependencies across the spectral domain.
Moreover, the horizontal distribution across channels is relatively diverse yet structured.
Some channels are consistently activated across most spectral bands, while others respond selectively.
This suggests that the network is learning to allocate different channels for modeling specific spectral contexts or frequency-dependent patterns.
Table~\ref{tab:globalcontext} compares different configurations of SCSS, including static statistics (Sta.), forward-only Mamba (F.W.), backward-only Mamba (B.W.), and bidirectional Mamba (Bi.D.).
The Sta. variant applies only channel-wise attention using global statistical priors, while the others incorporate dynamic Mamba scanning in one or both directions.
All configurations using channel attention outperform the CNN baseline, indicating that even simple spectral-wise modeling can provide significant gains, highlighting the importance of band-wise information.
Among the dynamic variants, F.W. and B.W. achieve similar performance, suggesting that both scanning directions are individually effective and contribute comparably.
The full Bi.D. configuration performs best across all metrics, while introducing only marginal additional computation.
These results confirm the effectiveness and efficiency of SCSS in enhancing spectral context representation.
These findings confirm that SCSS is not only structurally compact, but also capable of modeling the global, asymmetric, and non-uniform characteristics of hyperspectral spectra in a principled and interpretable manner.

\section{Conclusion}

In this paper, we revisit the overlooked issue of non-uniformity in hyperspectral image (HSI) representation and propose a principled solution named FairHyp.
Unlike prior approaches that apply uniform strategies or task-specific tricks, FairHyp disentangles and addresses spatial, spectral, and feature-level inconsistencies through three dedicated modules.
The RK4-SVA module effectively restores spatial coherence under heterogeneous resolutions via a fourth-order iterative approximation.
The S$^2$FairConv module promotes selective and efficient feature extraction by adapting to sparsity and structural diversity.
The SCSS module models long-range spectral dependencies with minimal spatial interference, achieving robust and fair spectral encoding.
By integrating these components into a unified, fairness-directed framework, FairHyp achieves state-of-the-art performance across diverse HSI tasks, including denoising, inpainting, super-resolution, and classification.
Moreover, extensive analysis confirms that each module independently contributes to performance while collectively avoiding trade-offs, enabling FairHyp to break the long-standing trilemma in HSI modeling.
We believe FairHyp offers a new perspective for designing adaptive and interpretable models under non-uniform data distributions, with broad implications beyond hyperspectral imaging.


\begin{IEEEbiography}[{\includegraphics[width=1in,height=1.25in,clip,keepaspectratio]{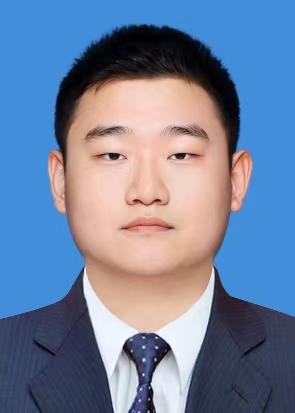}}]{Wuzhou Quan} 
received the B.E. degree in Electronic Information Engineering from Nanjing University of Posts and Telecommunications, China, in 2018, and the M.E. degree in Electronic Information from Shandong Technology and Business University, China, in 2024.
He is currently pursuing the Ph.D. degree in Computer Science and Technology at Nanjing University of Aeronautics and Astronautics, China.  
His research interests include computer vision, pattern recognition, and remote sensing.
\end{IEEEbiography}

\begin{IEEEbiography}[{\includegraphics[width=1in,height=1.25in,clip,keepaspectratio]{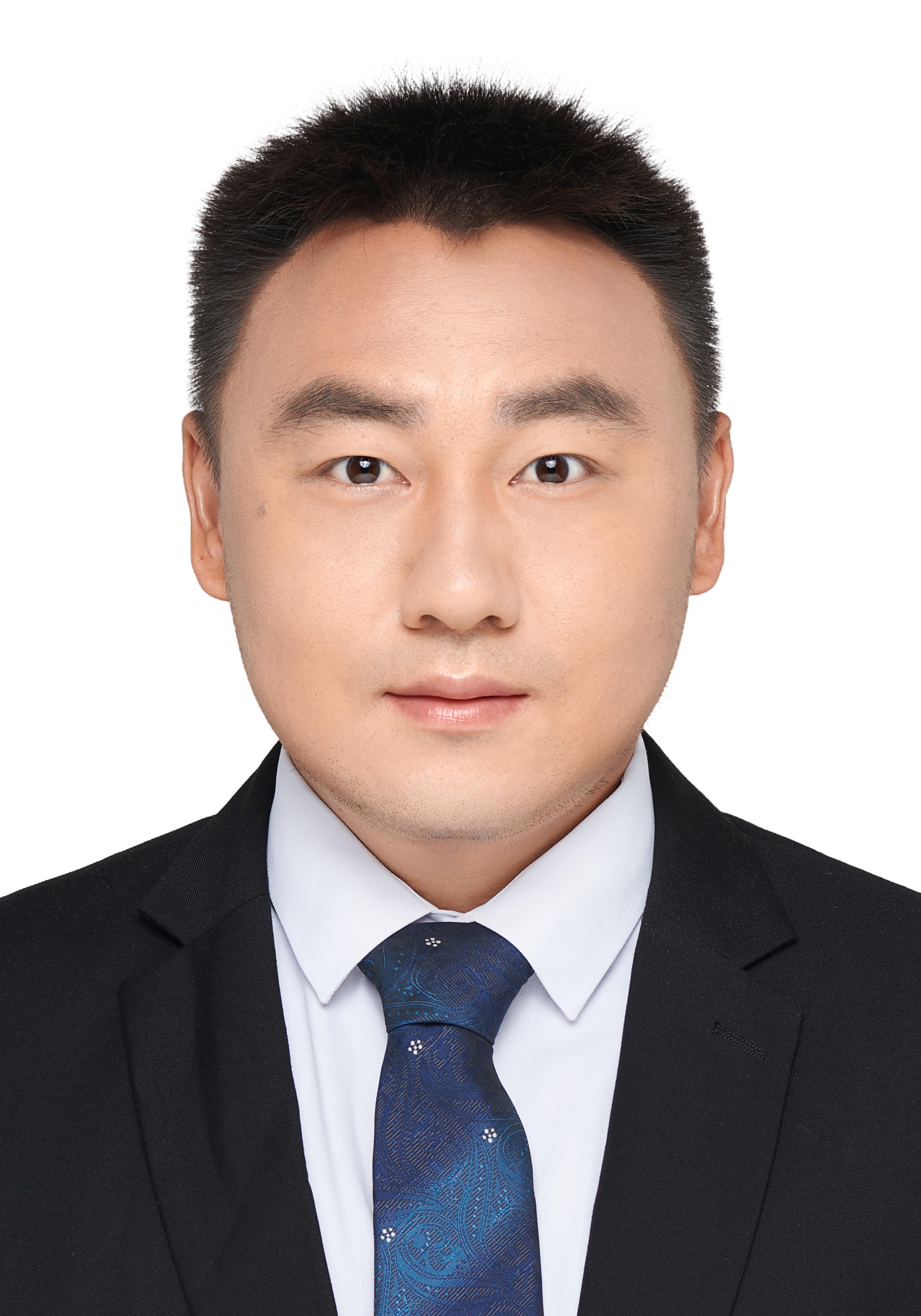}}]{Mingqiang Wei} (Senior Member, IEEE)
received his Ph.D degree (2014) in Computer Science and Engineering from the Chinese University of Hong Kong (CUHK). He is a professor at the School of Computer Science and Technology, Nanjing University of Aeronautics and Astronautics. He is now an Associate Editor for IEEE TIP and ACM TOMM, and was a Guest Editor for IEEE TMM. His research interests focus on 3D vision, and computer graphics.
\end{IEEEbiography}

\begin{IEEEbiography}[{\includegraphics[width=1in,height=1.25in,clip,keepaspectratio]{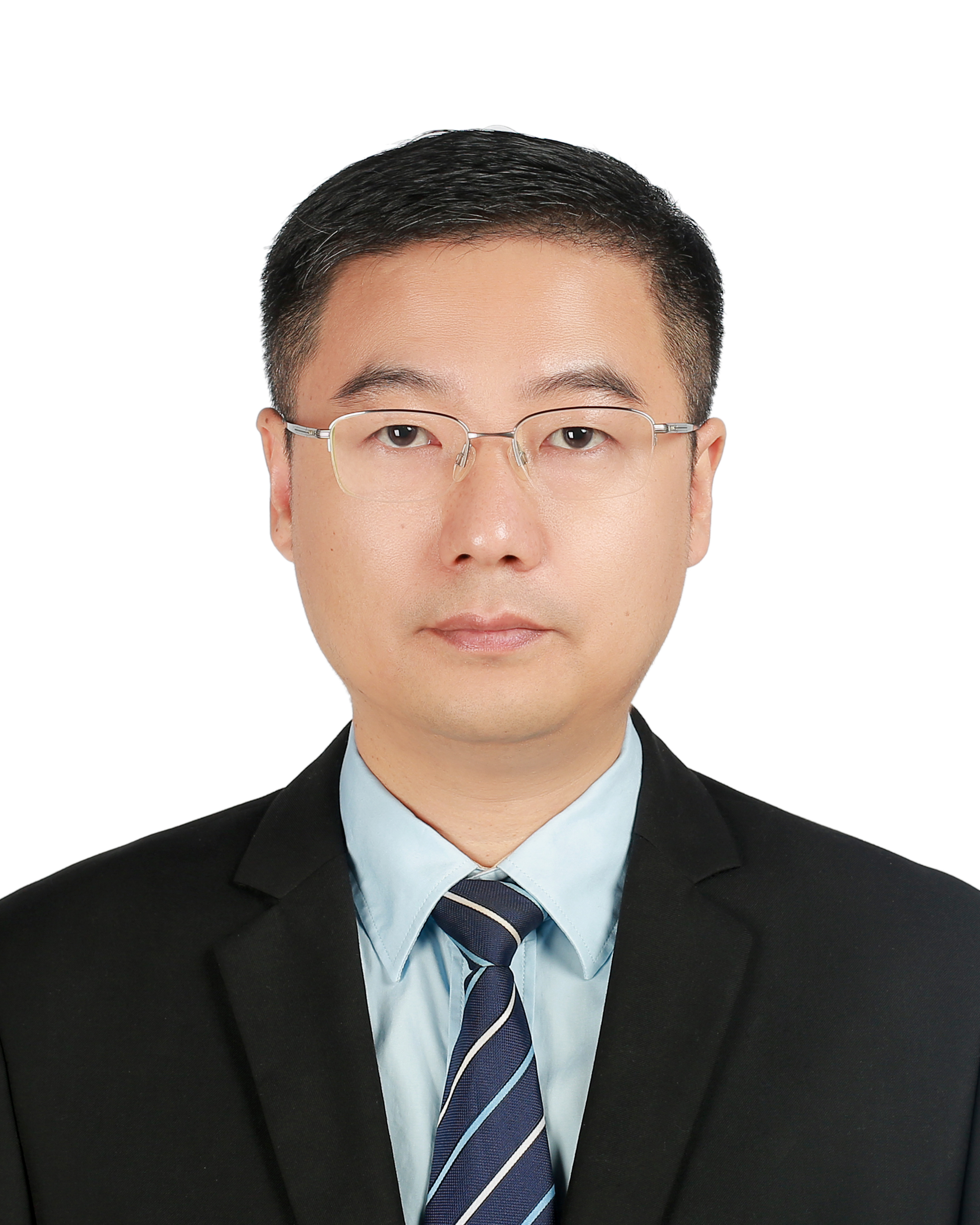}}]{Jinhui Tang} (Senior Member, IEEE)
eceived the BE and PhD degrees from the University of Science and Technology of China, Hefei, China, in 2003 and 2008, respectively.
He is currently a professor with the School of Computer Science and Engineering, Nanjing University of Science and Technology.
He received the best Student Paper Award in MMM 2016, and best paper awards in ACM MM 2007, PCM 2011, and ICIMCS 2011. He is also a member of the ACM.
\end{IEEEbiography}
\end{document}